%% file: main.tex
\documentclass{article}
\usepackage{iclr2027_conference,times}
\iclrfinalcopy   

\usepackage{microtype}
\input{math_commands.tex}

\usepackage{url}
\usepackage{amsmath,mathtools,amsfonts,amssymb}
\usepackage{xcolor}
\usepackage{graphicx}
\usepackage{caption}          
\usepackage{algorithm}
\usepackage{algpseudocode}
\usepackage[section]{placeins} 
\usepackage{float}            
\usepackage{booktabs}
\usepackage{enumitem}
\usepackage{pifont}
\usepackage{multicol,multirow}
\usepackage[T1]{fontenc}   
\usepackage{textcomp}      
\usepackage{listings}
\usepackage[varqu,varl,scaled=0.95]{zi4}
\usepackage{hyperref}
\hypersetup{
    colorlinks=true,       
    linkcolor=teal,        
    citecolor=teal,        
    urlcolor=magenta,      
    filecolor=teal,        
    bookmarks=true,        
    bookmarksnumbered=true,
    bookmarksopen=true,    
    pdfstartview=FitH      
}

\newcommand{\cmark}{\ding{51}}
\newcommand{\xmark}{\ding{55}}
\newcommand{\best}[1]{{\boldmath$#1$}}
\newcommand{\ci}[2]{\,{\scriptsize[#1,\,#2]}}
\definecolor{gaincol}{HTML}{1A7F37}
\definecolor{losscol}{HTML}{C0392B}

\newcommand{\gbase}{Gemma-4-12B}
\newcommand{\gdirect}{\textsc{OptiScribe}-12B-D}
\newcommand{\gladder}{\textsc{OptiScribe}-12B}
\newcommand{\qbase}{Qwen3.5-4B}
\newcommand{\qdirect}{\textsc{OptiScribe}-4B-D}
\newcommand{\qladder}{\textsc{OptiScribe}-4B}
\newcommand{\qladderxh}{\textsc{OptiScribe}-4B-xH}

\title{\centering Teaching LLMs to Generate Challenging MILP Instances via Solver Feedback}

\author{%
Jitin Singla$^\star$\\
IIT Roorkee\\
\texttt{jsingla@bt.iitr.ac.in}
\And
Parikshit Pareek$^\star$\\
IIT Roorkee\\
\texttt{pareek@ee.iitr.ac.in}
\AND
Pratik Jawanpuria\\
IIT Bombay\\
\texttt{pratik.jawanpuria@iitb.ac.in}
\And
Parag Singla\\
IIT Delhi\\
\texttt{parags@cse.iitd.ac.in}
}

\contribution{$^\star$Equal contribution and co-corresponding.\\
Authors acknowledge PARAM Pragya HPC facility IIT Delhi for computational resources used in this work.\\
JS and PP acknowledge Avathon Physical AI Lab, IIT Roorkee funding support from Avathon Inc. USA. PJ acknowledges the support of ANRF MATRICS grant. PS acknowledges support from IBM AI Horizon Network.\\
}

\begin{document}
\maketitle

\begin{abstract}
Generating optimization instances that are both feasible and computationally challenging is crucial for benchmarking solvers and training learning-based optimization algorithms. Existing non-LLM generators rely on seed instances or parameter tuning, resulting in high test-time computational cost, while existing LLM generators lack explicit hardness measures. Recent reinforcement learning methods with verifier feedback evaluate only binary correctness, which is misaligned with generating challenging problems. We note that an optimization solver reports the cost of solving at several stages of its pipeline, and leverage this to design a reward that scores both the solvability and the hardness of generated problems, measured by branch-and-bound nodes and post-cut relaxation gaps. Our key idea is a challenger-solver asymmetric self-play approach, where an LLM challenger generates progressively harder instances and the solver verifies feasibility and hardness, so no seed or training MILP instances are required. We fine-tune Gemma-4-12B and Qwen3.5-4B with GRPO and a size curriculum into OptiScribe-12B and OptiScribe-4B, which generate feasible yet challenging MILP problems from natural language instructions. On capacitated facility location and max-cut, OptiScribe-12B raises median SCIP search nodes by 1.7–5$\times$ and post-cut gaps by 1.1–1.7$\times$ over its base model and improves the feasibility rate on facility location by 9–19 points, while OptiScribe-4B raises median nodes by up to 15.6$\times$. The problems cover a wider difficulty range than public benchmarks of the same size, follow instructions on density and difficulty, and can tune solver settings for families that public libraries lack. These results indicate that optimization-specific rewards, used in self-play mode, can teach LLMs to generate high-difficulty optimization benchmarks. We will release our code and models publicly on acceptance.
\end{abstract}

\section{Introduction}
Generating hard test instances for a given class of optimization problems has long been recognized as a problem in its own right \citep{hooker1995testing, selman1996generating, smithmiles2015generating}. Hardness is usually known only after a solver has run, and instances produced by naive random generation are mostly easy \citep{cheeseman1991really, mitchell1992hard}. For mixed-integer linear programs (MILPs), which model scheduling, logistics, network design and energy systems, such instances are in practical demand. Solver developers need large and diverse instance sets for benchmarking and stress tests \citep{miplib2017}, and learning-augmented solvers need training instances from the right distribution \citep{gasse2019exact, bengio2021machine}. Yet public collections cover only a narrow slice of problem types and sizes, and at a given size their instances tend to be either trivial or very hard, with little in between (Section~\ref{sec:res-public}). Solver effort, the usual proxy for hardness, can be increased artificially by making instances larger, loosening big-$M$ constraints, or scaling coefficients badly, without making the problem structurally harder. A generator of hard MILP instances must therefore find where hardness lives without taking these shortcuts.

Although MILP instance generators have drawn growing attention \citep{bowly2019stress, geng2023g2milp, li2025milpevolve}, most do not learn what makes an instance hard. Their difficulty is set by hand, inherited from seed instances, or imposed by search after generation. Hand-written random generators, widely used in learning-based work, often yield small or easy instances \citep{prouvost2020ecole, huang2024distributional}. Seed-based methods edit given instances and aim to preserve their difficulty \citep{geng2023g2milp,guo2024acmmilp,liu2024milpstudio}, so their outputs stay close to the seeds and cannot be produced without them. LLM-based methods evolve or retrieve generator programs \citep{li2025milpevolve, yang2025milpretrieval}, but the LLM itself is never trained on hardness. Their difficulty comes from a solver-in-the-loop parameter search that must be rerun for each new class and size. A new family or size therefore needs new seeds, a new search or both.

Our starting point is that a MILP solver does more than return an answer. It also records how hard it worked, in branch-and-bound nodes explored and relaxation gap closed by cutting planes. We use this record to train a language model to write hard instances, which then needs no solver call at generation. Given a short prompt naming a MILP family and a target size, the model acts as a challenger and emits an instance in a compact index-set template. Training uses no instances beyond one formatting exemplar, which is never edited. A frozen branch-and-cut solver acts as verifier. The loop resembles challenger-solver self-play \citep{zhao2025absolutezero,huang2026rzero,dong2025stp}, where difficulty is defined by a co-trained solver whose skill shifts during training. Our verifier is fixed instead, so hardness is measured on one scale across checkpoints and base models.

We propose a reward designed to prevent the model from inflating solver effort without making the problem harder. A multiplicative validity gate requires a parseable, feasible instance with a finite optimum and rules out unbounded continuous variables, aggregated big-$M$ links, and extreme coefficient ranges. Hardness is scored from branch-and-bound nodes and the post-cut root gap, both read after root cuts \citep{achterberg2009scip}, so weaknesses that cutting planes repair earn little and difficulty confined to solvers with weaker cuts is largely not rewarded. A size term keeps instances near the requested size, so hardness cannot be bought with scale. Since hardness learned against one solver may still reflect its blind spots, we re-solve every instance at evaluation with HiGHS and the commercial solver Gurobi, neither of which enters training. We train with GRPO \citep{shao2024deepseekmath} and a curriculum that increases the variable count over time.

We train \gladder{} models from \texttt{Gemma-4-12B-it} \citep{gemma4technicalreport} and \qladder{} models from \texttt{Qwen3.5-4B} \citep{qwen38}. On capacitated facility location (CFL) and max-cut, both models generate harder instances than their base models. For \gladder{}, median SCIP branch-and-bound nodes rise by 1.7--5.0$\times$ on CFL and 1.9--4.5$\times$ on max-cut, median post-cut root gaps rise by 1.1--1.7$\times$, and the feasible rate on CFL improves by 9--19 points. The ordering of models holds at sizes larger than those seen in training and under the two held-out solvers. 
Our generator also produces instances across the difficulty range between the trivial and very hard instances that dominate public benchmarks of similar size. Overall, our contributions are as follows.
\begin{itemize}[nosep,leftmargin=0.5cm]
    \item \textbf{Learning hardness from a fixed verifier:} a seed-free self-learning loop in which an LLM challenger learns hardness against a frozen solver (Section~\ref{sec:methodology}).
    \item \textbf{A reward that resists false hardness:} a validity-gated hardness score read after presolve and root cuts, with size and structural-diversity terms (Section~\ref{sec:reward}).
    \item \textbf{Language-controlled generation:} The trained model generates instances directly from text, keeps the base model's ability to follow instructions on domain and difficulty, and shifts each instruction toward harder instances. Such instances can be used to tune a solver for a problem type that public libraries lack (Section \ref{sec:res-language}).
    \item \textbf{Released generators and evaluation:} \gladder{}, \qladder{}, and an evaluation protocol that checks hardness ordering across the used solvers.
\end{itemize}

\section{Background and Related Work}
\label{sec:related}

\subsection{Verifier-in-the-loop RL}
\label{subsec:verifierloop}

Recent work lets language models write their own training problems and score them with a verifier. Absolute Zero \citep{zhao2025absolutezero} trains one model to propose and solve code-reasoning tasks, rewarding tasks it solves only some of the time. R-Zero \citep{huang2026rzero} alternates a Challenger and a Solver from one base model and rewards questions on which the Solver matches its own majority vote about half the time, a label that weakens as questions get harder. STP \citep{dong2025stp} rewards conjectures the current prover proves only occasionally, with every proof checked by a formal verifier. In all three, difficulty is relative to a co-trained model, so the target moves during training and checkpoints share no fixed scale. Their verifiers also return only pass or fail, whereas a MILP solver reports how much effort a solve took. SIRL \citep{chen2025sirl} and ORLM \citep{huang2025orlm} use a solver to check formulations of user-provided problems, so they reward correctness rather than effort and create no new problems. We instead reward solver effort under a fixed verifier, which gives difficulty an absolute scale (Table~\ref{tab:positioning}).

\begin{table}[t]
\centering
\footnotesize
\setlength{\tabcolsep}{2.5pt}
\caption{Comparison with verifier-in-the-loop RL and MILP instance generators.}
\label{tab:positioning}
\begin{tabular}{@{}lllcclcc@{}}
\toprule
\multirow{2}{*}{Method} & \multirow{2}{*}{Artifact} & \multirow{2}{*}{Difficulty Signal} & Difficulty & Learns & Input at & Solver at & Text \\
 & & & Scale & Hardness & Generation & Inference & control \\
\midrule
\multicolumn{8}{@{}l}{\textit{Verifier-in-the-loop RL}}\\[2pt]
Absolute Zero & Code task & Solver success rate & Relative & \cmark & Past tasks & \cmark & \xmark \\
R-Zero & Question & Majority agreement & Relative & \cmark & None & \xmark & \xmark \\
STP & Conjecture & Prover pass rate & Relative & \cmark & Seed theorems & \xmark & \xmark \\
\midrule
\multicolumn{8}{@{}l}{\textit{MILP instance generators}}\\[2pt]
Random Gens. & Instance & None & Hand-set & \xmark & Parameters & \xmark & \xmark \\
G2MILP & Instance & None & Preserved & \xmark & Seed instances & \xmark & \xmark \\
ACM-MILP & Instance & None & Preserved & \xmark & Seed instances & \xmark & \xmark \\
MILP-StuDio & Instance & Solve time$^{a}$ & Preserved & \xmark & Seed instances & \xmark$^{a}$ & \xmark \\
MILP-Evolve & Code & Time, nodes, gap, size & Absolute & \xmark & Seed classes & \cmark & \xmark \\
MILP-Retrieval & Code & Similarity, time & Absolute & \xmark & Target instance & \cmark & \xmark \\
\textbf{\textsc{OptiScribe}} (Ours) & Instance & Validity, nodes, gap & Absolute & \cmark & Exemplar & \xmark & \cmark \\
\bottomrule
\multicolumn{8}{@{}p{\textwidth}@{}}{\footnotesize
\textit{Relative:} difficulty is set against co-trained model (drifts during training). \textit{Preserved}: reproduces supplied-seed difficulty. \textit{Hand-set}: fixed via manual parameters. \textit{Learns hardness}: generator trained on a difficulty signal. \textit{Text control}: hardness specified in natural language. $^{a}$MILP-StuDio generates instances without a solver; only its hard-instance study solves edits and keeps the slowest.}
\end{tabular}
\vspace{-1em}
\end{table}

\subsection{MILP instance generators}
\label{subsec:generators}

Most MILP instance generators do not learn what makes an instance hard. Hand-written generators for set cover, combinatorial auctions and facility location \citep{balas1980set, leytonbrown2000towards, cornuejols1991comparison} set difficulty through size and density and supply most training data in learning-based work \citep{gasse2019exact, prouvost2020ecole}, yet at moderate sizes solvers often close their instances quickly \citep{huang2024distributional}. Instance-space methods evolve instances toward target regions of a feature space \citep{smithmiles2015generating, bowly2019stress}. Seed-based methods edit existing instances by masked graph reconstruction \citep{geng2023g2milp}, grouped constraint modification \citep{guo2024acmmilp} or block manipulation \citep{liu2024milpstudio}. They aim to preserve the seed's difficulty and need seeds of the target family, and MILP-StuDio obtains hard instances only by keeping the slowest of repeated edits.
LLM-based methods write generator programs instead. MILP-Evolve \citep{li2025milpevolve} evolves generator code with a frozen LLM and keeps parameter settings whose instances hit solver-measured targets (size, solve time, node count, integrality gap). MILP-Retrieval \citep{yang2025milpretrieval} retrieves the library program closest to a target instance and tunes it by Bayesian optimization on solve time. In both, difficulty comes from searched parameters, the search reruns for each class and size, and solve-time targets vary with hardware and random seeds \citep{lodi2013variability}. None accepts a natural-language specification, so a user cannot simply ask for a denser graph or a harder instance. Our generator is trained directly against solver effort, needs no seed instance or solver call at generation, and takes its specification as text (Table~\ref{tab:positioning}).

\subsection{MILPs and solver effort signals}
\label{sec:solver-stages}

We consider MILPs of the form
\begin{equation}
\label{eq:milp}
\begin{aligned}
\mathcal{I}:\quad z^{\star} &= \min_{x_{C},\,x_{D}}\; c^{\top}[x_{C};x_{D}] \qquad \text{s.t.}\quad A\,[x_{C};x_{D}] \le b,\qquad \ell \le [x_{C};x_{D}] \le u,\\
&x_{C}\in\mathbb{R}^{d},\; x_{D}\in\mathbb{Z}^{\,n-d},\; c\in\mathbb{R}^{n},\; A\in\mathbb{R}^{m\times n},\; b\in\mathbb{R}^{m},
\end{aligned}
\end{equation}
where $n$ is the number of variables, $m$ the number of constraints, and $d$ the number of continuous variables. A discrete variable with bounds $0$ and $1$ is binary. Bounds may be infinite, and we set $z^{\star}=+\infty$ if $\mathcal{I}$ is infeasible and $z^{\star}=-\infty$ if it is unbounded.

A branch-and-cut solver works in four phases, and each reports a quantity that could serve as a reward. \emph{Presolve} shrinks $\mathcal{I}$ to an equivalent instance by propagating bounds, fixing forced variables, and tightening coefficients. A large reduction signals redundancy rather than difficulty. The \emph{root relaxation} drops integrality and gives a bound $z_{\mathrm{LP}} \le z^{\star}$, and the difference $z^{\star} - z_{\mathrm{LP}}$ is the integrality gap of the formulation. The \emph{root cutting loop} adds valid inequalities that tighten this bound to $z_{\mathrm{cut}}$, with $z_{\mathrm{LP}} \le z_{\mathrm{cut}} \le z^{\star}$. \emph{Search} then closes the remaining gap $z^{\star} - z_{\mathrm{cut}}$ using $N(\mathcal{I})$ branch-and-bound nodes. In  Section~\ref{sec:reward}, we design a  reward function using these quantities.

\section{Proposed Seed-Free Asymmetric Self-Play Approach}
\label{sec:methodology}
Our goal is an LLM that, given a MILP family and a target size, writes a valid instance that is hard to solve. With no library of hard instances to learn from, the model must discover hardness itself. It proposes instances, a solver attempts them, and the model is rewarded by the solver's effort, learning over training which choices make the solver work harder.

\subsection{Challenger and verifier}
\label{sec:selfplay}

We build on challenger-solver self-play, but unlike prior work we do not train the solver. The LLM policy $\pi_\theta$ acts as the challenger and proposes instances, while a branch-and-cut solver $\mathcal{S}$ with a fixed configuration acts as the verifier, checking feasibility, proving optimality, and reporting the search effort required. Because only the challenger learns, the game is asymmetric.
Training is also seed-free. Prompts only mention problem family, target size, output format, and one worked example at a different size. We never use example to target hardness, or instance corpus, or solver call at inference time. Appendix~\ref{app:prompt} gives the training prompt.

\subsection{A solver-effort reward}
\label{sec:reward}

We build the reward from the quantities the solver reports (Section~\ref{sec:solver-stages}). As discussed earlier, it must separate an instance that the solver closes at the root from one that forces a long search, without being inflated by formulations that only look hard.

\paragraph{Where to read the reward:}
\label{sec:where-read}
Rewarding the raw gap $|z^{\star}-z_{\mathrm{LP}}|$ would also reward weak but equivalent formulations. For example, $\sum_j x_j \le Mz$ with a large $M$ lets the relaxation take a fractional $z$ and inflates the gap, although the decision problem is unchanged. Disaggregated links $x_j\le u_j z$ largely remove this, and presolve and cuts often do so anyway. We therefore read the reward after presolve and the root cutting loop, so that it measures the difficulty that remains after the solver's own repairs. Section~\ref{sec:ablation-precut} gives supporting evidence that difficulty read before cuts can be specific to one solver. We define reward for each completion $o_i$, which parsed into an instance $\mathcal{I}_i$ as:
\begin{equation}
R_i \;=\; V(\mathcal{I}_i) \cdot
\Bigl[\,w_{H}\,H(\mathcal{I}_i) + w_{v}\,r_{\mathrm{var}}(\mathcal{I}_i) + w_{d}\,r_{\mathrm{div}}(\mathcal{I}_i)\,\Bigr],
\label{eq:reward}
\end{equation}
where $V$ is a validity gate, $H$ the hardness, $r_{\mathrm{var}}$ a size term, and $r_{\mathrm{div}}$ a diversity term, with weights $(w_H, w_v, w_d)$. The gate $V(\mathcal{I}_i)\in\{0,1\}$ equals 1 only if the instance parses, matches the requested family, and the solver finds a feasible solution with a finite optimum. It also requires bounded continuous variables, a bounded coefficient range, and no big-$M$ constraint that ties many continuous variables to one binary, which rules out apparent hardness that vanishes under presolve or comes from ill-conditioning (Appendix~\ref{app:reward-gate}). For a valid instance, hardness combines the post-cut gap with the size of the search tree,
\begin{equation}
H(\mathcal{I}_i) = 0.25\, r_{\mathrm{cut}}(\mathcal{I}_i)\,+\,0.75\, r_{\mathrm{node}}(\mathcal{I}_i),
\label{eq:hard}
\end{equation}
\begin{equation}
\small
g_{\mathrm{cut}}(\mathcal{I}_i) = \frac{\bigl|z^{\star}_i - z_{\mathrm{cut},i}\bigr|}{\tau\,|z^{\star}_i|},
\,\,
r_{\mathrm{cut}}(\mathcal{I}_i) = \mathrm{clip}\bigl(g_{\mathrm{cut}}(\mathcal{I}_i),0,1\bigr),
\,\,
r_{\mathrm{node}}(\mathcal{I}_i) = \mathrm{clip}\!\left(\frac{\log N(\mathcal{I}_i)}{\log N_{\mathrm{ref}}},0,1\right).
\label{eq:reward-node}
\end{equation}
With $\tau=0.10$, relative gaps are resolved up to $10\%$ and larger ones are capped, and the absolute value makes $r_{\mathrm{cut}}$ independent of the optimization sense. Node counts are heavy-tailed, so $r_{\mathrm{node}}$ is log-scaled with $N_{\mathrm{ref}}=5{,}000$ and is zero for an instance solved at the root ($N=1$). Because a large post-cut gap does not guarantee a long search, the node term gets the larger weight and the gap term serves for shaping. Since the easiest way to raise $r_{\mathrm{node}}$ is to write larger instances, the size term rewards closeness to the requested size,
\begin{equation}
r_{\mathrm{var}}(\mathcal{I}_i) \;=\;\exp\!\Bigl(-\alpha\,\dfrac{\lvert n_{i} - n^{\star}\rvert}{n^{\star}}\Bigr),
\label{eq:reward-var}
\end{equation}
where $n_i$ is the realized variable count, $n^{\star}$ the midpoint of the requested bracket, and $\alpha=3$. The term is soft, so an instance outside the bracket loses at most $w_v$. The diversity term $r_{\mathrm{div}}$ ranks each valid instance by how rare its structural fingerprint is within its group, which discourages the policy from collapsing onto one structure. It averages exactly $0.5$ in every group and therefore only redistributes advantage among valid instances (Appendix~\ref{app:reward-div}). Appendix~\ref{app:reward-example} gives a worked example of the complete reward.
We also train \qladderxh{}, that drops the diversity term and moves its weight to hardness.

\begin{figure}[t]
    \centering
    \includegraphics[width=\linewidth]{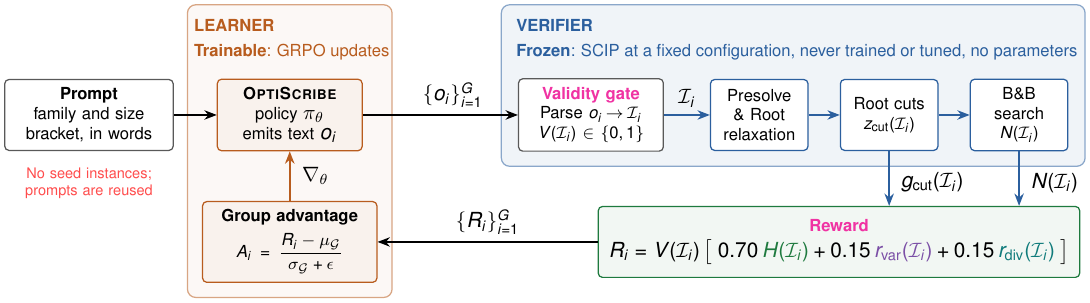}
    \vspace{-1.3em}
    \caption{\textbf{Seed-free asymmetric self-play with trainable challenger and frozen verifier.} From a prompt naming a family and a size bracket, the policy generates $G{=}64$ instances $\mathcal I_i$, which SCIP solves with a fixed configuration. The verifier returns the validity gate $V$, post-cut gap $g_{\mathrm{cut}}$ and node count $N$, to compute hardness $H$. $H$ is then combined with size $r_{\mathrm{var}}$ and diversity $r_{\mathrm{div}}$ rewards for GRPO-based Challender update. The untrained verifier keeps difficulty on a fixed scale.}
    \label{fig:diagram}
\end{figure}

\subsection{Instance representation}
\label{sec:instanceTemplate}

Each instance is written in a compact index-set template rather than as an explicit coefficient matrix. The template has a header (family and optimization sense), a structure block (index sets, variable types and bounds, an objective, and constraint families quantified over the sets), and a data block of numeric arrays. A deterministic parser expands this text into an explicit MILP. Because the structure block does not grow with the instance, token cost grows with the number of parameters rather than the number of constraints (Appendix~\ref{app:index-set}). Token caps are set per size bracket from the length distribution of complete instances, and a completion that exceeds its cap fails to parse and receives zero reward.

\subsection{Training loop}
\label{sec:loop}

Figure~\ref{fig:diagram} illustrates a single training iteration. Given a prompt $p$, the policy samples a group of $G$ completions $\{o_i\}_{i=1}^{G}$. We parse each completion into an instance $\mathcal{I}_i$ (Section~\ref{sec:instanceTemplate}), solve it with $\mathcal{S}$, and compute its reward $R_i$ (\ref{eq:reward}). We fine-tune the policy with LoRA \citep{hu2022lora} using group relative policy optimization (GRPO) \citep{shao2024deepseekmath}, which takes the within-group standardized reward as the advantage,
$A_i = (R_i - \operatorname{mean}_{j} R_j) /(\operatorname{std}_{j} R_j + \epsilon).$ A completion is therefore reinforced only when it outperforms its peers for the same prompt, for example by yielding a harder instance, or a valid one where its peers fail. We apply the clipped GRPO update with a KL penalty that keeps the policy close to the base model (Appendix~\ref{app:impl}). Each group is drawn from a single problem family and size bracket, and families alternate across steps, so instances are never compared across families.

\section{Empirical Results and Discussion}
\label{sec:results}

We train two base LLMs, \gbase{} and \qbase{}, on three standard MILP families that differ in structure: capacitated facility location (CFL) has linking constraints, linearized Max-Cut is a graph problem, and multiple knapsack is a packing problem. We evaluate the generated instances along two axes, their gain in hardness over the untrained base model and their standing against public benchmarks of similar size, and then examine robustness, design choices and uses. Our experiments address six questions. \textbf{1)} Do trained models generate harder instances than their base models (Section~\ref{sec:res-main})? \textbf{2)} How do these instances compare with public benchmarks of similar size (Section~\ref{sec:res-public})? \textbf{3)} Does the added difficulty hold under solvers not used in training (Section~\ref{sec:solver-ablation})? \textbf{4)} How do reward design choices, in particular where the reward is read, affect the results (Section~\ref{sec:ablation-precut})? \textbf{5)} Can the trained model still be steered with language (Section~\ref{sec:res-language})? \textbf{6)} Can generated instances be used to tune solver settings (Section~\ref{sec:res-tune})?

\textbf{Experimental setup:} We train each base model in two ways. The \emph{curriculum} models, \gladder{} and \qladder{}, train on three size brackets in succession, $76$--$110$ (B0), $111$--$170$ (B1) and $171$--$225$ (B2) variables, with each stage initialized from the previous one. The \emph{direct} models, \gdirect{} and \qdirect{}, train on B2 alone with the same reward and hyperparameters. We evaluate on B1, B2, B3 ($226$--$350$) and B4 ($351$--$500$), so B3 and B4 test size extrapolation. \gbase{} is our primary model, and the smaller \qbase{} replicates the main results and carries the node-reference and reward-weight ablations. Unless stated otherwise, the reward weights are $(w_H, w_v, w_d) = (0.70, 0.15, 0.15)$, and SCIP runs on one thread with a $50{,}000$-node cap (Appendix~\ref{sec:training-config}). All models receive identical prompts (Appendix~\ref{app:prompt}) with matched sampling seeds, so comparisons are paired. We report parse rate, feasible rate, branch-and-bound nodes and the post-root-cut gap $|z^\star - z_{\mathrm{cut}}|/|z^\star|$, where more nodes and larger gaps mean more solver effort. We avoid solve time, which depends on the machine, except in Section~\ref{sec:res-tune}. Appendices~\ref{app:impl} and~\ref{app:geometries} detail the curriculum, geometry choice, hyperparameters and measurement conventions.

\subsection{Trained models generate harder instances than their base models}
\label{sec:res-main}

Table~\ref{tab:main-f5-g2-split} reports paired comparisons for Gemma. In every bracket, \gladder{} generates the hardest CFL and Max-Cut instances, raising median nodes over \gbase{} by $1.7$--$5.0\times$ (CFL) and $1.9$--$4.5\times$ (Max-Cut) and median post-cut gaps by $1.4$--$1.7\times$ and $1.1$--$1.3\times$. The gains persist at the unseen sizes B3--B4 ($2.4\times$ nodes on CFL, $2.4\times$/$1.9\times$ on Max-Cut), and the central $95\%$ range of node counts is $1.7$--$5.2\times$ wider, covering a broader span of difficulty. On CFL, parse and feasible rates also rise by $9$--$15$ and $9$--$19$ points. On Max-Cut, where \gbase{} already parses nearly everything, the only notable drop ($76.6\%$ vs.\ $98.8\%$ parsed at B4) comes mostly from unfinished completions, as every parsed \gladder{} instance is feasible.

Direct training at the target size fails for Gemma. \gdirect{} became unstable as more completions hit the token cap and scored zero, and we stopped it after $47$ of $61$ updates (Appendix Figure~\ref{fig:training_curve}), with nodes and gaps near or below those of \gbase{}. Multiple knapsack stays easy for all models, with median gaps of about $0.5\%$ (Appendix~\ref{app:res-knapsack}).

\textbf{A second base model:} To test transfer across LLMs, we repeat the analysis on \qbase{} (Table~\ref{tab:scip-base-qwen-gains-main}). \qladder{} matches or improves parsing (by up to $17$ points on CFL), raises median nodes by $2.9$--$11.6\times$ on Max-Cut and $1.6$--$15.6\times$ on CFL, and raises median gaps by $9.7$--$11.1$ and $0.8$--$2.1$ points, respectively. Unlike Gemma, Qwen trains stably at the target bracket (\qdirect{} improves on \qbase{} in $21$ of $24$ cells), yet within the training brackets \qladder{} still beats \qdirect{} in $10$ of $12$ comparisons, with one tie. The curriculum is thus necessary for Gemma and beneficial for Qwen. Qwen models also tend to generate harder instances than Gemma (Appendix Figures~\ref{fig:features_f5} and~\ref{fig:features_g2}), partly because \qbase{} is already stronger on Max-Cut, so the difference reflects the base models rather than the framework.

\begin{table}[t]
\small
\centering
\caption{\textbf{Main results for CFL and Max-Cut under SCIP.} $256$ instances per arm and bracket. Nodes is the median node count and Gap the median post-root-cut gap ($\times10^{-3}$), both over feasible instances, with the $2.5$th and $97.5$th percentiles in brackets. Bold marks the best value.}
\label{tab:main-f5-g2-split}
{\footnotesize\setlength{\tabcolsep}{2.3pt}%
\resizebox{\linewidth}{!}{%
\begin{tabular}{l@{\hspace{6pt}}lcccc@{\hspace{6pt}}cccc}
\toprule
& & \multicolumn{4}{c}{Capacitated Facility Location (CFL)} &
    \multicolumn{4}{c}{Max-Cut} \\
\cmidrule(lr){3-6}\cmidrule(lr){7-10}
\rotatebox[origin=c]{90}{Bracket} & Arm & Parse\,$\uparrow$ & Feas\,$\uparrow$ & Nodes\,$\uparrow$ & Gap\,$\uparrow$ & Parse\,$\uparrow$ & Feas\,$\uparrow$ & Nodes\,$\uparrow$ & Gap\,$\uparrow$ \\
\midrule
\multirow{3}{*}{B1}
 & \gbase{}   & 78.9 & 72.7 & 22\ci{1}{456} & 18\ci{1}{63} & \textbf{100.0} & 85.9 & 41\ci{15}{91} & 258\ci{182}{314} \\
 & \gdirect{} & 84.8 & 82.4 & 21\ci{1}{388} & 18\ci{1}{55} & 99.6 & \textbf{89.1} & 41\ci{15}{90} & 260\ci{190}{315} \\
 & \gladder{} & \textbf{94.1} & \textbf{91.8} & \textbf{109}\ci{2}{755} & \textbf{30}\ci{2}{74} & \textbf{100.0} & \textbf{89.1} & \textbf{142}\ci{29}{297} & \textbf{338}\ci{237}{392} \\
\cmidrule(lr){1-10}
\multirow{3}{*}{B2}
 & \gbase{}   & 88.3 & 86.7 & 224\ci{1}{1873} & 22\ci{4}{63} & 99.6 & 91.8 & 81\ci{29}{182} & 298\ci{237}{354} \\
 & \gdirect{} & 85.5 & 85.2 & 170\ci{1}{1587} & 24\ci{2}{65} & 99.6 & \textbf{96.9} & 85\ci{31}{193} & 300\ci{238}{355} \\
 & \gladder{} & \textbf{99.6} & \textbf{96.5} & \textbf{391}\ci{1}{3539} & \textbf{35}\ci{3}{82} & 99.6 & 89.8 & \textbf{364}\ci{103}{821} & \textbf{396}\ci{316}{456} \\
\cmidrule(lr){1-10}
\multirow{3}{*}{B3}
 & \gbase{}   & 86.3 & 82.0 & 293\ci{1}{6272} & 20\ci{2}{52} & \textbf{100.0} & 95.3 & 423\ci{201}{744} & 395\ci{332}{446} \\
 & \gdirect{} & 87.5 & 86.7 & 211\ci{1}{5591} & 20\ci{2}{62} & 98.8 & 95.7 & 369\ci{148}{676} & 377\ci{317}{436} \\
 & \gladder{} & \textbf{98.0} & \textbf{94.9} & \textbf{710}\ci{2}{18296} & \textbf{27}\ci{2}{70} & 99.6 & \textbf{96.1} & \textbf{1002}\ci{510}{1770} & \textbf{465}\ci{405}{529} \\
\cmidrule(lr){1-10}
\multirow{3}{*}{B4}
 & \gbase{}   & 76.6 & 75.8 & 420\ci{1}{20671} & 16\ci{1}{64} & \textbf{98.8} & \textbf{97.3} & 1350\ci{823}{2180} & 485\ci{426}{537} \\
 & \gdirect{} & 69.1 & 68.4 & 194\ci{0}{10658} & 14\ci{1}{40} & 82.4 & 80.5 & 1096\ci{477}{1801} & 456\ci{379}{515} \\
 & \gladder{} & \textbf{85.5} & \textbf{84.4} & \textbf{1014}\ci{2}{34277} & \textbf{25}\ci{2}{58} & 76.6 & 76.6 & \textbf{2578}\ci{1011}{8122} & \textbf{546}\ci{467}{623} \\
\bottomrule
\end{tabular}%
}%
}
\end{table}

\begin{table}[t]
\centering
\caption{Qwen arms against untrained \qbase{} under SCIP. Parsing and median post-root-cut gap are differences in percentage points (pp). Nodes is the ratio of median node counts. $128$ completions per model and cell, feasible instances only. \qladderxh{} raises the hardness weight and drops the diversity term. Bold marks the largest value.}
\label{tab:scip-base-qwen-gains-main}
\footnotesize
\setlength{\tabcolsep}{3.3pt}
\renewcommand{\arraystretch}{1.05}
\resizebox{\linewidth}{!}{%
\begin{tabular}{@{}llccc@{\hspace{8pt}}ccc@{}}
\toprule
& & \multicolumn{3}{c}{Max-Cut} & \multicolumn{3}{c}{Capacitated facility location} \\
\cmidrule(lr){3-5}\cmidrule(lr){6-8}
Bracket & Arm & Parsing (pp) & Nodes & Gap (pp) & Parsing (pp) & Nodes & Gap (pp) \\
\midrule
\multirow{3}{*}{111--170}
 & \qdirect{}           & $+4.69$      & $4.96\times$       & $+10.61$      & $+14.84$      & $2.23\times$        & $-0.27$ \\
 & \qladder{}           & $+2.34$      & $5.08\times$       & $+10.74$      & \best{+17.19} & $12.23\times$       & $+1.14$ \\
 & \qladderxh{} & \best{+9.38} & \best{7.22\times}  & \best{+13.40} & \best{+17.19} & \best{26.68\times}  & \best{+2.75} \\
\cmidrule(lr){1-8}
\multirow{3}{*}{171--225}
 & \qdirect{}           & $-0.78$      & $4.74\times$       & $+10.40$      & \best{+1.56}  & $6.93\times$        & $+1.13$ \\
 & \qladder{}           & \best{+0.78} & $5.26\times$       & $+11.02$      & \best{+1.56}  & $15.63\times$       & $+1.45$ \\
 & \qladderxh{} & \best{+0.78} & \best{6.56\times}  & \best{+14.14} & \best{+1.56}  & \best{26.51\times}  & \best{+1.81} \\
\cmidrule(lr){1-8}
\multirow{3}{*}{226--350}
 & \qdirect{}           & $-1.56$      & $3.38\times$       & $+12.37$      & $+8.59$       & $12.69\times$       & $+0.54$ \\
 & \qladder{}           & $0.00$       & $2.85\times$       & $+9.66$       & \best{+9.38}  & $7.64\times$        & \best{+2.08} \\
 & \qladderxh{} & \best{+5.47} & \best{3.82\times}  & \best{+13.38} & \best{+9.38}  & \best{31.51\times}  & $+0.53$ \\
\cmidrule(lr){1-8}
\multirow{3}{*}{351--500}
 & \qdirect{}           & $+6.25$      & $16.63\times$      & $+12.62$      & $+3.12$       & $2.76\times$        & $+0.19$ \\
 & \qladder{}           & \best{+7.81} & $11.58\times$      & $+11.11$      & $+4.69$       & $1.61\times$        & \best{+0.83} \\
 & \qladderxh{} & \best{+7.81} & \best{28.83\times} & \best{+14.35} & \best{+5.47}  & \best{11.14\times}  & $+0.37$ \\
\bottomrule
\end{tabular}%
}
\end{table}

\subsection{Generated instances fill gaps in public benchmarks}
\label{sec:res-public}
\begin{figure}[t]
    \centering
    \vspace{-1em}
    \includegraphics[width=\linewidth]{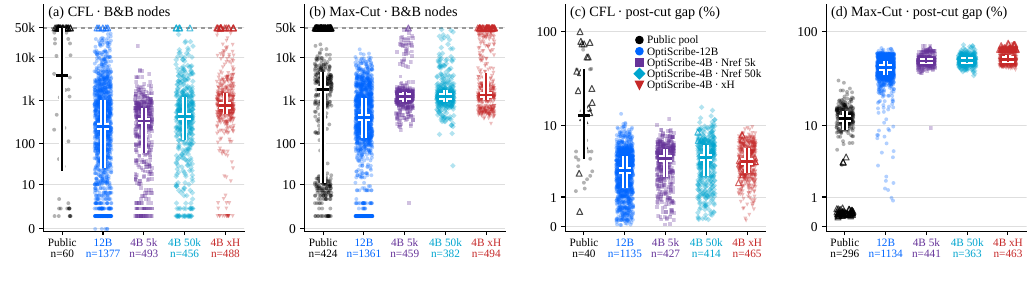}
    \vspace{-1.5em}
    \caption{\textbf{Generated instances against public benchmarks, $111$--$500$ variables.} Public mixed-integer instances against generated CFL (a, c) and public pure-integer instances against generated Max-Cut (b, d). (a, b) SCIP nodes, (c, d) post-root-cut gap in percent. Points are single instances, all weighted equally. Crossbars are medians and vertical bars the interquartile range. Triangles mark instances that hit the $50{,}000$-node cap (dashed line), and their gaps are measured against the best incumbent.}
    \label{fig:public}
\end{figure}

We pool public instances with $111$--$500$ variables from eight sources, including MIPLIB 2010/2017, MILP-Evolve, and D-MIPLIB (full list and citations in Appendix~\ref{app:public-datasets}). We compare public mixed-integer instances against our CFL pools, and public pure-integer instances against our Max-Cut pools, using \gladder{}, two \qladder{} runs ($N_{\mathrm{ref}}=5{,}000$ and $50{,}000$, Section~\ref{sec:ablation-precut}), and \qladderxh{} (Figure~\ref{fig:public}). Public coverage in this range is sparse and largely bimodal. Among the $60$ mixed-integer instances, most are either trivial ($\approx 10$ nodes) or nearly hit the $50{,}000$-node cap, while the pure-integer set adds only a narrow band around $1{,}000$--$10{,}000$ nodes. Our generated pools span the full difficulty spectrum, with several instances reaching the cap, and \qladderxh{} reaches the cap about twice as often as the other pools. Public mixed-integer instances solved to optimality typically end with gaps below $\approx 6\%$, similar to our CFL pools, whereas our Max-Cut pools retain much larger post-cut gaps ($41.9$--$49.9\%$ vs.\ $11.9\%$). Standard synthetic generators are easier still at this scale. Their pooled median instance is solved at the root and none needs more than $189$ nodes, against median node counts of $347$ on CFL and $634$ on Max-Cut for \gladder{} (Appendix~\ref{app:res-ecole}). Since our generator produces instances of a requested family and size on demand, we use it for tuning in Section~\ref{sec:res-tune}.

\subsection{The hardness ordering holds under held-out solvers}
\label{sec:solver-ablation}
\begin{figure}
    \centering
    \includegraphics[width=\linewidth]{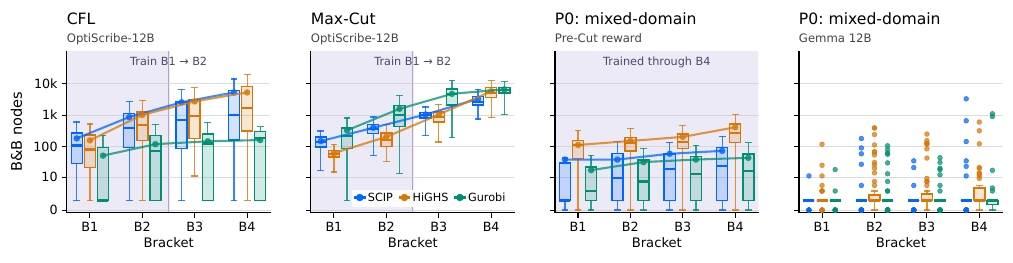}
    \vspace{-2em}
    \caption{\textbf{Node counts under three solvers.} (a, b) \gladder{} CFL and Max-Cut instances under SCIP (training solver), HiGHS and Gurobi, with training brackets shaded. (c) A model trained on prompt P0 with a pre-cut reward and HiGHS in the loop, through all four brackets. (d) Untrained \gbase{} on P0. Boxes show median and interquartile range, dots are outliers, lines join per-bracket medians. Log vertical axis.}
    \label{fig:solver_ablation}
\end{figure}

Since the reward comes from SCIP, the model could learn instances that are hard only for SCIP. We re-solve all instances with HiGHS and Gurobi~13.0.3, neither used in training (Figure~\ref{fig:solver_ablation}a,b). For \gladder{}, median nodes grow with size under every solver. HiGHS follows SCIP's trend, while Gurobi needs fewer nodes on CFL and more on Max-Cut. The model ranking is unchanged: \gladder{} has the highest median in every bracket except CFL B1 under Gurobi, where all medians are root-solved (Appendix~\ref{app:res-solvers}). Whenever both prove optimality, SCIP and Gurobi agree on the optimal value within tolerance, so the difficulty is not an artifact of numerical error.

\subsection{Reward design: readout point, node reference and weights}
\label{sec:ablation-precut}
We ablate three reward choices. \emph{Readout point.} Section~\ref{sec:where-read} discussed that a reward read before root cuts can be inflated by weak formulations. A model trained with a pre-cut reward and HiGHS in the loop, on the mixed-domain prompt P0 (Appendix~\ref{app:promptP0}), illustrates this. Its instances tie binaries to continuous sums through single loose big-$M$ rows and leave $67$--$95\%$ of variables without an upper bound, so the LP relaxation opens facilities almost for free. 
Tightening $M$ to a valid value left HiGHS node counts unchanged on all $1{,}715$ instances. SCIP's root cuts, by contrast, repair the relaxation, and disabling them raises SCIP's node count by a median factor of $8$--$10$. SCIP and Gurobi therefore solve one-third to over half of these instances at the root, and HiGHS at most $6\%$ (Figure~\ref{fig:solver_ablation} c,d; Appendix Figure~\ref{fig:solver_triviality}), which is why we read the reward after root cuts. 
\emph{Raising the node reference} $N_{\mathrm{ref}}$, where the node term in \eqref{eq:reward-node} saturates, to $50{,}000$ for Qwen changes little inside the training brackets (Appendix Figure~\ref{fig:nref}). Outside them it raises CFL node gains at B3/B4 ($24.5\times$/$5.7\times$ vs.\ $7.6\times$/$1.6\times$) but lowers Max-Cut at B4 ($6.6\times$ vs.\ $11.6\times$) and costs over $10$ points of parse rate. \emph{The hardness weight} acts as a dial. Raising it and dropping the diversity term (\qladderxh{}) gives the highest node count in every cell of Table~\ref{tab:scip-base-qwen-gains-main}. That is $3.8$--$28.8\times$ (Max-Cut) and $11.1$--$31.5\times$ (CFL) over \qbase{}, or $1.2$--$2.5\times$ and $1.7$--$6.9\times$ over \qladder{}, with no parse-rate loss but tighter clustering in Figure~\ref{fig:public}, i.e.\ less diversity.

\subsection{Language control is preserved on Max-Cut}
\label{sec:res-language}
\begin{figure}
    \centering
    \includegraphics[width=\linewidth]{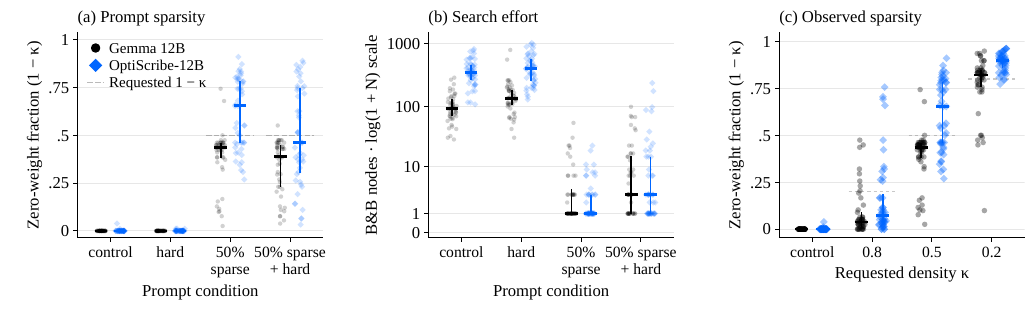}
    \caption{\textbf{Language control over Max-Cut.} Bracket $171$--$225$, $48$ instances per model/prompt, with SCIP. (a) Share of zero-weight vertex pairs, $1-\kappa$, under each prompt. Dashed lines are requested values. (b) Branch-and-bound nodes. (c) Realized zero-weight share vs. requested density. Points are single completions, horizontal bars are medians and vertical bars span the interquartile range.}
    \label{fig:prompt_control}
\end{figure}

An LLM generator should do what the prompt says. For Max-Cut at $171$--$225$ variables, we append one unseen sentence to a fixed prompt, requesting a harder instance, a sparser graph at target edge density $\kappa$, or both (Appendix Table~\ref{tab:sentences}), and sample $48$ completions per sentence from \gbase{} and \gladder{} (Figure~\ref{fig:prompt_control}). \emph{Both models follow the density instructions.} For requested densities $0.8/0.5/0.2$, median realized densities are $0.96/0.56/0.18$ for \gbase{} and $0.93/0.35/0.10$ for \gladder{}, which overshoots but keeps the order. ``Harder'' raises node counts from $92$ to $135$ for \gbase{} and from $344$ to $400$ for \gladder{}, so training raises the level at which a sentence lands ($3.7\times$ and $3.0\times$ over \gbase{}) without amplifying its marginal effect. On CFL, \gladder{} does not track the analogous density request, though its instances still become easier (Appendix Figure~\ref{fig:sparsity_effect}). \emph{The effect is bounded by problem structure.} Our Max-Cut LP bound equals the total edge weight (Appendix~\ref{app:index-set-maxcut}), so sparser graphs raise the maximum cut's share of it. At densities $0.5$ and $0.2$ the median share rises from about $0.6$ to $0.86$--$1.00$, the bound becomes nearly tight and SCIP often closes at the root, and even ``harder'' with a sparse request reaches a median of only $3$ nodes. New requirements can thus be stated as unseen sentences, and the trained model follows them within what the family's mathematics allows.

\subsection{Generated instances support solver tuning}
\label{sec:res-tune}
Solver tuning works best on instances that match the target workload, but public collections often miss the right family or size. We test whether \gladder{} can fill that gap by tuning SCIP on $40$ generated instances with $226$--$350$ variables and evaluating the chosen setting on held-out instances.  \emph{In-distribution utility:} MILP-Evolve has no CFL instances at this size, so we generated them with \gladder{} and tested on $24$ held-out generated instances.  The best setting caps aggregation-based cutting planes at five rounds at the root (instead of no round limit), reducing mean solve time by $16.0\%$ (95\% bootstrap CI $6.3$--$25.5\%$) with identical optimal values.
\emph{Out-of-distribution transfer:} We selected among $10$ SCIP configurations on generated Max-Cut instances and applied the winning configuration unchanged to $40$ same-size MILP-Evolve combinatorial-auction instances. The wining configuration disables cutting planes and solves $35/40$ instances within $10$\,s, compared to $14/40$ with default settings. It also cuts mean PAR2\footnote{Penalized average runtime, the mean solve time with each unsolved instance counted at twice the time limit \citep{froleyks2021satcomp}.} from $15.15$ to $7.50$\,s (paired difference $-7.64$\,s, 95\% bootstrap CI $-9.43$ to $-5.88$\,s). Tuning directly on MILP-Evolve selects the same setting. 
Generated instances thus recover the setting chosen on the target library, and they allow tuning for a family and size that the library lacks.

\section{Conclusions and Limitations}
We showed that solver effort after root cuts is a usable reward for teaching LLMs to write hard MILP instances. With a frozen verifier and no seed instances or instance corpus, \gladder{} and \qladder{} generate harder CFL and Max-Cut instances than their base models at trained and larger sizes, and the ranking holds for two held-out solvers. The instances cover difficulty levels that public benchmarks of similar size leave sparse, and stay controllable through natural-language requests such as density and difficulty. They are also useful in practice: settings tuned on them match those tuned on MILP-Evolve, and they allow tuning for a family and size that public libraries lack.

Our study has limitations. It covers three families, and multiple knapsack stays easy for every model. Generated instances currently have at most $500$ variables. We train each arm once and tune on a small set of instances. Broader families, larger problems, and wider evaluations are left for future work.

\FloatBarrier

\section*{Author Contributions}
\noindent J.S. and P.P. developed the methodology, implemented the system, ran the experiments, and performed the analysis. All other authors contributed to study design and interpretation of results, provided feedback throughout the project, and participated in writing, reviewing, and approving the final manuscript. P.S. arranged the computational resources required for the project.


\bibliography{main}
\bibliographystyle{plainnat}

\newpage
\appendix

\begin{center}
\Large 
    \textbf{Appendix: Teaching LLMs to Generate Challenging MILP Instances via Solver Feedback}
\end{center}

\lstdefinestyle{promptstyle}{
  basicstyle=\ttfamily\scriptsize,
  breaklines=true,
  breakindent=0pt,
  postbreak=\mbox{$\hookrightarrow$\space},
  columns=fullflexible,
  keepspaces=true,
  upquote=true,
  frame=single,
  framesep=4pt,
  xleftmargin=4pt,
  escapeinside={(*@}{@*)},
  literate={,}{{,}\allowbreak}1
}

\section{The generation prompt}
\label{app:prompt}
\subsection{Scope and delivery}
Both models \gladder{} and \qladder{} are trained with the \emph{same} prompt. The models only differ only in the underlying base model and in the chat template applied by that model's tokenizer. Concretely, the rendered prompt is passed in as a \textbf{single user message, with no system
prompt}. Each training row has the form:
\begin{center}
{\small\texttt{\{"role": "user", "content": [\{"type": "text", "text": <rendered prompt>\}]\}}}
\end{center}
The trainer then applies the model's chat template. As a result, the only input-level
difference between the two models is the template wrapper (Gemma's
\texttt{<start\_of\_turn>user} versus Qwen's \texttt{<|im\_start|>user}), not the prompt body.

We render one prompt per training example using a function family
$f \in $ \{Capacitated Facility Location, Max-cut, Multiple knapsack\}, the curriculum bracket $b$, and an in-context exemplar.

\subsection{Template}

Every prompt is the concatenation of seven blocks in fixed order:

\begin{enumerate}\itemsep0pt
  \item a one-line task statement,
  \item \textbf{FAMILY / MANDATORY MATHEMATICS}: the per-family structural mandate (Listing~\ref{lst:mandates}),
  \item \textbf{SIZE}: the variable-count interval $[\ell_b, u_b]$ for the bracket plus a natural-language geometry hint (Table~\ref{tab:cellparams}),
  \item \textbf{FEASIBILITY PROCEDURE}: a per-family construction recipe with one bracket-dependent numeric constant substituted in (Table~\ref{tab:cellparams}),
  \item \textbf{WHAT YOU MUST NOT DO}: four shared prohibitions (Listing~\ref{lst:shared});
  \item \textbf{PRECEDENCE}: the conflict rule (Listing~\ref{lst:shared});
  \item \textbf{EXAMPLE}: one complete in-context exemplar, followed by the strict output-format instruction.
\end{enumerate}

\begin{lstlisting}[style=promptstyle,caption={Template skeleton. Braced tokens are
substituted per training example.},label={lst:skeleton}]
Write ONE mixed-integer linear program in the format shown by the example below.

{MANDATE[family]}

SIZE: the problem must have between {lo} and {hi} variables in total. Aim for {geometry_hint}.
A two-index family x[A,B] contributes |A| x |B| variables; a one-index family x[A] contributes
|A|. Choose the set sizes so the total lands in range.

FEASIBILITY PROCEDURE (follow it; do not merely aim at the outcome):
  {procedure[family] with {caphint} substituted}
  A model with no feasible solution is worth nothing, and so is one whose optimum is obvious.

{BANS}

{PRECEDENCE}

EXAMPLE of the required format and mathematics:

{exemplar}

Now write a NEW problem of the same family at the size requested above.
Output exactly ONE problem.

OUTPUT FORMAT -- this is strict:
  Begin with the MILP header line. End with the DATA block. Stop immediately after the closing
  brace of the DATA block.
  Write NO comments, NO explanation, NO reasoning, NO checks, and no text before or after the
  model. Every line you write must be part of the model itself.
\end{lstlisting}

\subsection{Per-family mandates}

\begin{lstlisting}[style=promptstyle,caption={The three family mandates, verbatim.},label={lst:mandates}]
=== Capacitated facility location ===
FAMILY: tight facility location.
MANDATORY MATHEMATICS (the names are yours; the structure is not):
  - CONTINUOUS flow variables indexed by TWO sets, x[F,D], each with an explicit finite
    upper bound
  - BINARY open variables indexed by one set, y[F]
  - a demand row per customer:   for d in D: sum f in F: x[f,d] >= dem[d]
  - a PER-PAIR link:             for f in F, d in D: x[f,d] - dem[d]*y[f] <= 0
  - a capacity row per depot WITHOUT any binary: for f in F: sum d in D: x[f,d] <= cap[f]
  - a minimisation objective over flow cost plus opening cost

=== Multiple knapsack ===
FAMILY: multiple knapsack (packing items into containers).
MANDATORY MATHEMATICS (the names are yours; the structure is not):
  - one family of BINARY variables indexed by TWO sets, x[I,K] (item, container)
  - each item used at most once:  for i in I: sum k in K: x[i,k] <= 1
    this row MUST be `<=`, never `=` -- with `=` a tight capacity makes the problem
    impossible instead of hard
  - a weighted capacity row per container: for k in K: sum i in I: w[i]*x[i,k] <= cap[k]
  - a MAXIMISATION objective over a two-index profit, p[i,k]

=== Maximum cut, linearised ===
FAMILY: maximum cut on a graph, written in the standard linearised form.
MANDATORY MATHEMATICS (the names are yours; the structure is not):
  - BINARY node variables indexed by one set, x[V]
  - BINARY edge variables indexed by two sets, y[V,V]
  - BOTH of these rows, as a pair, over the same (i,j):
        for i in V, j in V: y[i,j] - x[i] - x[j] <= 0
        for i in V, j in V: y[i,j] + x[i] + x[j] <= 2
    Neither row alone models a cut: the first alone lets y be 0 everywhere, the second
    alone lets y be 1 everywhere.
  - a MAXIMISATION objective over a two-index edge weight, w[i,j]
\end{lstlisting}

\subsection{Shared blocks}

\begin{lstlisting}[style=promptstyle,caption={Prohibitions and precedence rule, identical for
every family and bracket.},label={lst:shared}]
WHAT YOU MUST NOT DO:
  1. Do not leave any continuous variable without an explicit finite upper bound.
     `var x[F,D] 0 inf continuous` is invalid. State a real number.
  2. Do not write an aggregated single-binary link -- one binary switching off a SUM of two or
     more continuous columns, as in `sum d in D: x[f,d] - 1000*y[f] <= 0`. A per-pair link,
     `x[f,d] - dem[d]*y[f] <= 0`, is the correct form.
  3. Do not use coefficients spanning more than four orders of magnitude. Keep every nonzero
     within a factor of 10,000 of every other.
  4. Do not omit any mandatory row above, and do not add a row that makes another one
     non-binding.

PRECEDENCE, when these instructions appear to conflict:
  the validity gates dominate these instructions, and these instructions dominate the exemplar.
  The exemplar shows ONE way to satisfy the mandate. It is not the only way, and where it
  differs from the instructions above, the instructions win.
\end{lstlisting}

\subsection{Per-cell parameters}

Only two things vary with the curriculum bracket: the geometry hint in the \textbf{SIZE} line
and one numeric constant in the \textbf{FEASIBILITY PROCEDURE}. The procedure sentences are
otherwise fixed per family:

\begin{itemize}\itemsep0pt
  \item \textbf{Facility Location}: ``Pick every demand \texttt{dem[d]} as an integer from 5 to 20. Then set EVERY \texttt{cap[f]} to \emph{c}, and every opening cost \texttt{fopen[f]} to an integer from 120 to 260, large enough that opening one more depot is a real decision, not free.''
  \item \textbf{Multiple knapsack}: ``Pick every weight \texttt{w[i]} as an integer from 10 to 30 and every profit \texttt{p[i,k]} as an integer from 20 to 60. Then set EVERY \texttt{cap[k]} to \emph{c}. Do not compute it from the weights, just use that number.''
  \item \textbf{Max-cut}: ``Set \texttt{w[i,i]} = 0 on the whole diagonal. For every pair $i$ below $j$, set \texttt{w[i,j]} to an integer from 1 to 20. Every single pair, leaving none at zero. Then mirror it: \texttt{w[j,i]} = \texttt{w[i,j]}.'' (No bracket-dependent constant.)
\end{itemize}

Every geometry hint in Table~\ref{tab:cellparams} lands inside its bracket (CFL: $|F||D| + |F|$; multiple knapsack: $|I||K|$; Max-Cut: $|V|^2 + |V|$). For CFL the constant $c$ also guarantees feasibility for any demand draw: in every cell $|F|\,c \ge 20\,|D|$, so total capacity exceeds the largest possible total demand.

\begin{table}[h]
\centering
\small
\caption{Geometry hint and capacity constant $c$ by family and bracket. Multiple knapsack is not trained at
76--110: no screened geometry for that family branched at that size, so the cell is excluded
from the curriculum rather than assigned a hint.}
\label{tab:cellparams}
\begin{tabular}{llll}
\toprule
Family & Bracket & Geometry hint (\textbf{SIZE} line) & $c$ \\
\midrule
CFL & 76--110   & about 8 depots and 12 customers  & 41 \\
CFL & 111--170  & about 10 depots and 15 customers & 41 \\
CFL & 171--225  & about 14 depots and 13 customers & 26 \\
CFL & 226--350  & about 16 depots and 17 customers & 29 \\
CFL & 351--500  & about 20 depots and 20 customers & 28 \\
\midrule
Multiple knapsack & 76--110   & \multicolumn{2}{l}{\emph{not trained at this bracket}} \\
Multiple knapsack & 111--170  & about 42 items and 4 containers  & 105 \\
Multiple knapsack & 171--225  & about 48 items and 4 containers  & 120 \\
Multiple knapsack & 226--350  & about 48 items and 6 containers  & 80 \\
Multiple knapsack & 351--500  & about 48 items and 8 containers  & 60 \\
\midrule
Max-cut & 76--110   & about 9 nodes  & --- \\
Max-cut & 111--170  & about 12 nodes & --- \\
Max-cut & 171--225  & about 13 nodes & --- \\
Max-cut & 226--350  & about 16 nodes & --- \\
Max-cut & 351--500  & about 20 nodes & --- \\
\bottomrule
\end{tabular}
\end{table}

\subsection{In-context exemplar}

Each prompt carries exactly one exemplar, drawn uniformly at random from a fixed pool of three per family. Exemplars are rendered at a \emph{fixed} geometry that does not depend on the bracket: CFL's pool has 104, 150 and 104 variables, Multiple knapsack's and Max-cut's pools are at 192 and 182 variables respectively. The exemplar therefore lies inside the requested interval only in a few brackets. In lower brackets it is larger than requested, and in higher brackets smaller. Its data are also not re-drawn with the bracket's constant. The exemplar in Listing~\ref{lst:full} has capacities 22--25, whereas the procedure asks for 41. The \textbf{PRECEDENCE} block exists to resolve these conflicts in favour of the instructions. Provenance comments are stripped before embedding so that comment lines are not demonstrated as part of the output format.

\subsection{A complete rendered prompt}

Listing~\ref{lst:full} is one prompt exactly as the model receives it (CFL at bracket
111--170), before the chat template is applied.

\begin{lstlisting}[style=promptstyle,caption={A complete training prompt: family CFL, bracket
111--170. The \texttt{DATA} line is a single long line in the original and is wrapped here.},label={lst:full}]
Write ONE mixed-integer linear program in the format shown by the example below.

FAMILY: tight facility location.
MANDATORY MATHEMATICS (the names are yours; the structure is not):
  - CONTINUOUS flow variables indexed by TWO sets, x[F,D], each with an explicit finite
    upper bound
  - BINARY open variables indexed by one set, y[F]
  - a demand row per customer:   for d in D: sum f in F: x[f,d] >= dem[d]
  - a PER-PAIR link:             for f in F, d in D: x[f,d] - dem[d]*y[f] <= 0
  - a capacity row per depot WITHOUT any binary: for f in F: sum d in D: x[f,d] <= cap[f]
  - a minimisation objective over flow cost plus opening cost

SIZE: the problem must have between 111 and 170 variables in total. Aim for about 10 depots and 15 customers.
A two-index family x[A,B] contributes |A| x |B| variables; a one-index family x[A] contributes
|A|. Choose the set sizes so the total lands in range.

FEASIBILITY PROCEDURE (follow it; do not merely aim at the outcome):
  Pick every demand dem[d] as an integer from 5 to 20. Then set EVERY cap[f] to 41, and every opening cost fopen[f] to an integer from 120 to 260 -- large enough that opening one more depot is a real decision, not free.
  A model with no feasible solution is worth nothing, and so is one whose optimum is obvious.

WHAT YOU MUST NOT DO:
  1. Do not leave any continuous variable without an explicit finite upper bound.
     `var x[F,D] 0 inf continuous` is invalid. State a real number.
  2. Do not write an aggregated single-binary link -- one binary switching off a SUM of two or
     more continuous columns, as in `sum d in D: x[f,d] - 1000*y[f] <= 0`. A per-pair link,
     `x[f,d] - dem[d]*y[f] <= 0`, is the correct form.
  3. Do not use coefficients spanning more than four orders of magnitude. Keep every nonzero
     within a factor of 10,000 of every other.
  4. Do not omit any mandatory row above, and do not add a row that makes another one
     non-binding.

PRECEDENCE, when these instructions appear to conflict:
  the validity gates dominate these instructions, and these instructions dominate the exemplar.
  The exemplar shows ONE way to satisfy the mandate. It is not the only way, and where it
  differs from the instructions above, the instructions win.

EXAMPLE of the required format and mathematics:

MILP tight_facility min
set F 10
set D 14
par cost[F,D]
par fopen[F]
par dem[D]
par cap[F]
var x[F,D] 0 20 continuous
var y[F] 0 1 binary
obj min sum f in F, d in D: cost[f,d]*x[f,d] + sum f in F: fopen[f]*y[f]
con demand: for d in D: sum f in F: x[f,d] >= dem[d]
con link: for f in F, d in D: x[f,d] - dem[d]*y[f] <= 0
con cap: for f in F: sum d in D: x[f,d] <= cap[f]
DATA: {"cost":[[6,12,1,10,5,1,15,9,9,8,13,14,4,7],[1,8,10,13,14,12,11,1,11,8,10,12,11,15],[1,15,7,11,9,14,6,8,11,13,6,14,1,4],[13,14,8,11,9,2,7,2,2,12,12,15,11,6],[12,3,11,10,8,11,11,2,7,8,7,15,14,1],[6,9,9,14,3,13,11,1,15,7,3,15,3,2],[6,6,4,4,9,5,6,6,15,8,4,4,5,15],[11,3,7,5,11,8,2,12,14,7,14,14,7,11],[11,6,11,11,11,14,14,6,11,3,12,4,14,8],[4,2,8,6,1,13,9,13,14,5,14,10,9,13]],"fopen":[198,172,240,154,202,242,255,175,173,165],"dem":[12,14,12,11,11,20,8,13,11,15,18,14,20,20],"cap":[22,23,25,23,23,23,25,23,25,23]}

Now write a NEW problem of the same family at the size requested above.
Output exactly ONE problem.

OUTPUT FORMAT (*@\textemdash@*) this is strict:
  Begin with the MILP header line. End with the DATA block. Stop immediately after the closing
  brace of the DATA block.
  Write NO comments, NO explanation, NO reasoning, NO checks, and no text before or after the
  model. Every line you write must be part of the model itself.
\end{lstlisting}


\section{The Solver-Based GRPO Reward Function}
\label{app:reward}
In this section, we present the details of two reward components that were reffered to Appendix section for more details: Validity gate and Structural diversity.

\subsection{The validity gate}
\label{app:reward-gate}

The validity gate for instance $\mathcal{I}_i$ generated while training is $V(\mathcal{I}_i) = \prod_{k=1}^{6}\mathbf{1}[g_k(\mathcal{I}_i)]$ with the following conditions.

\begin{itemize}\itemsep2pt
\item[$g_1$] \textbf{Parse:} The emitted instance expands under the fixed grammar, with all index sets declared, all referenced parameters present and all rows well formed.
\item[$g_2$] \textbf{Well posed:} The solver returns a feasible and a finite optimal value within the node cap and the wall-clock limit. Infeasible models, unbounded models and killed solves all fail here.
\item[$g_3$] \textbf{Bounded variables:} Every continuous variable carries a finite upper bound. Without this gate the policy inflates relaxation gaps with unbounded variables which is a route to apparent hardness that vanishes under standard pre-processing.
\item[$g_4$] \textbf{No aggregated single-binary link:} No constraint gates a sum of continuous variables through a single binary with one large coefficient. The banned pattern couples many continuous columns to one switch, which we found makes instances easier for the solver, not harder, while inflating superficial structure.
\item[$g_5$] \textbf{Coefficient range:} The ratio of the largest to the smallest nonzero magnitude over the constraint matrix and objective is at most $10^{4}$. This blocks hardness manufactured from numerical ill-conditioning, which is an artefact of tolerances rather than of combinatorial structure.
\item[$g_6$] \textbf{Family compliance:} Declared index sets, variable types and row families match the family named in the prompt.
\end{itemize}

\subsection{Structural diversity}
\label{app:reward-div}

GRPO only compares completions \emph{within} the same sampling group, so we also measure structural diversity within a group. We compress each instance to a simple fingerprint $\phi(I)$. For each constraint row we record: (i) the row sense, (ii) the number of nonzeros, (iii) how many of those nonzeros sit in integer columns, and (iv) the set of coefficient signs. The fingerprint is the multiset of these row ``types''. By design it forgets variable names, coefficient magnitudes, and row order. That is deliberate. Two instances that differ only cosmetically should map to the same fingerprint, which otherwise the diversity term would reward mere renaming, which is exactly the failure mode it is meant to avoid.

Let $\mathcal{G}=\{I_1,\dots,I_m\}$ denote the \emph{valid} completions in a sampling group, i.e., those that pass the validity gate (invalid completions are dropped and receive score $0$). Hence $m$ is at most the nominal group size. For each completion $I_k$, define
\begin{equation}
  c_k \;=\; \bigl\lvert\{\,j : \phi(I_j)=\phi(I_k)\,\}\bigr\rvert ,
  \qquad
  \rho_k \;=\; -c_k ,
\end{equation}
where $c_k$ is the number of group members that share $I_k$'s fingerprint, and $\rho_k$ is a simple ``rarity'' score (less frequent fingerprints yield larger $\rho_k$). We then turn these rarities into tie-averaged ranks and rescale to $[0,1]$:
\begin{equation}
  r_k \;=\; L_k + \frac{T_k-1}{2},
  \qquad
  L_k = \bigl\lvert\{\,j : c_j > c_k\,\}\bigr\rvert,
  \qquad
  T_k = \bigl\lvert\{\,j : c_j = c_k\,\}\bigr\rvert,
\end{equation}
\begin{equation}
  r_{\mathrm{div}}(k) \;=\;
  \begin{cases}
    \dfrac{r_k}{m-1}, & m > 1,\\[2mm]
    0.5, & m = 1 .
  \end{cases}
\end{equation}
Note that $L_k$ and $T_k$ count \emph{completions} (not distinct fingerprints), and $T_k$ includes $k$ itself. By construction, $r_{\mathrm{div}}(k)\in[0,1]$.

This definition is intentionally neutral in the two degenerate regimes. If a group collapses to a single fingerprint, then every completion has $c_j=m$, the ranks tie, and $r_{\mathrm{div}}\equiv 0.5$. If every fingerprint is unique, then every completion has $c_j=1$, the ranks tie again, and $r_{\mathrm{div}}\equiv 0.5$ as well. In both cases the term is constant within the group and therefore cannot create spurious advantage differences. The statistic becomes informative only when a group contains a mix of common and rare structures. Because GRPO normalizes rewards within each group, only within-group variation in $r_{\mathrm{div}}$ affects the update. The term is zero-sum around $0.5$ and it lowers the advantage of over-represented structures and raises that of under-represented ones, discouraging the policy from concentrating mass on a single structure.

In our experiments, each group is drawn from a single family, and families alternate between optimization steps. This ensures the advantage never pits (say) knapsack against max-cut, so $r_{\mathrm{div}}$ always measures \emph{within-family} structural variety.

\subsection{Worked example}
\label{app:reward-example}

Consider a completion in a group of $64$ at the bracket $171$--$225$, so $n^{\star} = 198$, which parses, passes every gate, is detected as compliant, and solves to optimality with $196$ variables, $1207$ nodes, $z^{\star} = z_{\mathrm{IP}} = 4820$, and root dual bound $z_{\mathrm{cut}} = 4510$. Then
\begin{align}
r_{\mathrm{nodes}} &= \log 1207 / \log 5000 = 0.833, \nonumber\\
r_{\mathrm{cut}} &= 310/(0.10\times4820) = 0.643, \nonumber\\
H &= 0.75 \cdot 0.833 + 0.25 \cdot 0.643 = 0.786, \nonumber\\
r_{\mathrm{var}} &= \exp(-3 \cdot 2/198) = 0.970, \nonumber
\end{align}
With a within-group diversity rank of $0.42$,
\begin{equation}
R = 0.70 \cdot 0.786 + 0.15 \cdot 0.970 + 0.15 \cdot 0.42 = 0.759 .
\end{equation}


\section{Compact Index-set Template}
\label{app:index-set}

Each instance is written in two parts. The \emph{structure block} describes the model once, its index sets, parameter tables, variable families, a single objective, and constraint families quantified over the sets. The \emph{data block} is a JSON object that provides the values for each parameter table as nested arrays. We never write the coefficient matrix $A$ in \eqref{eq:milp} and instead, a deterministic parser reconstructs it by expanding each family over its index sets. Table~\ref{tab:app-grammar} summarizes the grammar statements.

\begin{table}[H]
\centering
\small
\caption{\textbf{Statements of the index-set template:} Each line of the structure block is one statement. $S$ and $T$ are declared index sets, $i \in S$ and $j \in T$ are their running indices, and \texttt{op} is one of \texttt{<=}, \texttt{>=}, \texttt{=}. Indices are bound by \texttt{sum} and \texttt{for} clauses; in a \texttt{var} statement, the running index of each set is its lowercase name.}
\label{tab:app-grammar}
\begin{tabular}{@{}p{0.40\linewidth}p{0.54\linewidth}@{}}
\toprule
Statement & Meaning \\
\midrule
\texttt{MILP <family> <min|max>} & Three-token header with problem class, family name and optimization sense. \\
\texttt{set S <size>} & Index set $S = \{1,\dots,|S|\}$. \\
\texttt{par p[S,T]} & Parameter table $p \in \mathbb{R}^{|S|\times|T|}$, values given in the data block. \\
\texttt{var v[S,T] [<lb> <ub>] <type>} & Variable family $v_{ij}$, $(i,j)\in S\times T$, with type \texttt{continuous}, \texttt{integer} or \texttt{binary}. A bound is a number or a parameter entry indexed by the running indices of the family, e.g.\ \texttt{dem[d]}. Bounds are omitted for \texttt{binary}, which fixes them to $[0,1]$. \\
\texttt{obj <min|max> <sum>{ + <sum>}} & Objective, a sum of terms such as \texttt{sum i in S, j in T: c[i,j]*v[i,j]}. \\
\texttt{con <name>: for i in S\{, j in T\}: <lhs> op <rhs>} & Constraint family with one row for every element of the product of the listed sets. \\
\texttt{DATA: \char123"p": [[...],...], ...\char125} & Numerical values of every declared parameter, nested row-major over its index sets. \\
\bottomrule
\end{tabular}
\end{table}

\textbf{Expansion:} The parser applies five rules. First, each variable family is instantiated row-major over its index sets, and the columns of \eqref{eq:milp} are ordered by family in declaration order. Second, each constraint family produces one row per element of its \texttt{for} sets, again row-major. Third, a product of a parameter entry and a variable, such as \texttt{dem[d]*z[f]}, becomes the numerical coefficient of that variable, so every row stays linear. 
Fourth, variable terms on the right-hand side are moved to the left with their sign reversed. Fifth, repeated occurrences of the same variable in a row are merged by summing their coefficients, so that, e.g., \texttt{x[i] + x[j]} with $i=j$ becomes $2x_i$.

\subsection{Capacitated facility location}
\label{app:index-set-cfl}

Let $F$ be the set of facilities and $D$ the set of customers. A binary variable $z_f$ opens facility $f$ at fixed cost $o_f$, and a continuous variable $x_{fd}$ is the amount customer $d$ receives from facility $f$ at unit cost $c_{fd}$. With demand $\delta_d$ and capacity $\kappa_f$,
\begin{align}
\min_{x,\,z}\quad & \sum_{f\in F} o_f\, z_f + \sum_{f\in F}\sum_{d\in D} c_{fd}\, x_{fd} \label{eq:app-cfl}\\
\text{s.t.}\quad & \textstyle\sum_{f\in F} x_{fd} \ge \delta_d, && d\in D, \tag{demand}\\
& x_{fd} - \delta_d\, z_f \le 0, && f\in F,\ d\in D, \tag{link}\\
& \textstyle\sum_{d\in D} x_{fd} \le \kappa_f, && f\in F, \tag{capacity}\\
& 0 \le x_{fd} \le \delta_d,\quad z_f\in\{0,1\}, && f\in F,\ d\in D. \nonumber
\end{align}
The instance has $n = |F|(1+|D|)$ variables, of which $|F|$ are binary, and $m = |D| + |F||D| + |F|$ rows. The link is written per facility and customer pair with coefficient equal to the column bound $\delta_d$, and the capacity row carries no binary. This is the disaggregated form discussed in Section~\ref{sec:methodology}, so the instance passes gates $g_3$ and $g_4$ of Appendix~\ref{app:reward-gate} by construction. The same model in the template reads

\begin{footnotesize}
\begin{verbatim}
MILP facility_location min
set F 2
set D 3
par fopen[F]
par cap[F]
par dem[D]
par cost[F,D]
var x[F,D] 0 dem[d] continuous
var z[F] binary
obj min sum f in F: fopen[f]*z[f] + sum f in F, d in D: cost[f,d]*x[f,d]
con demand: for d in D: sum f in F: x[f,d] >= dem[d]
con link: for f in F, d in D: x[f,d] - dem[d]*z[f] <= 0
con capacity: for f in F: sum d in D: x[f,d] <= cap[f]
DATA: {"fopen":[20,70], "cap":[28,32], "dem":[6,18,6],
       "cost":[[1,2,2],[2,4,2]]}
\end{verbatim}
\end{footnotesize}

\noindent Changing $|F|$ or $|D|$ changes the two \texttt{set} lines and the array lengths in \texttt{DATA}, and nothing else. For this data the columns are $x_{11},x_{12},x_{13},x_{21},x_{22},x_{23}$ followed by $z_1,z_2$, and the expansion gives $n=8$ variables and $m=11$ rows,
\begin{align*}
&x_{11}+x_{21}\ge 6, \qquad x_{12}+x_{22}\ge 18, \qquad x_{13}+x_{23}\ge 6, \\
&x_{11}-6z_1\le 0,\qquad x_{12}-18z_1\le 0,\qquad x_{13}-6z_1\le 0,\\
&x_{21}-6z_2\le 0,\qquad x_{22}-18z_2\le 0,\qquad x_{23}-6z_2\le 0,\\
&x_{11}+x_{12}+x_{13}\le 28, \qquad x_{21}+x_{22}+x_{23}\le 32,
\end{align*}
with objective $20z_1+70z_2+x_{11}+2x_{12}+2x_{13}+2x_{21}+4x_{22}+2x_{23}$ and bounds $x_{fd}\in[0,\delta_d]$.
Total demand is $30$, which exceeds $\kappa_1=28$, so facility~1 cannot serve all customers alone. Opening only facility~2 is feasible but costs $166$, whereas opening both costs $v^\star=144$, attained for instance by $x_{11}=6$, $x_{12}=18$, $x_{23}=6$. The optimum is not unique, since customer~3 costs the same from either facility and any split with $x_{13}\le 4$ is also optimal. The LP relaxation instead sets $z_1 = 14/15$ and $z_2 = 1/15$, just enough to cover the two units that facility~1 cannot supply, and attains $v_{\mathrm{LP}}=1202/15\approx 80.13$. Most of the relaxation gap comes from paying the fixed cost of the second facility fractionally.

\subsection{Max-cut linearization}
\label{app:index-set-maxcut}

Let $V$ be the vertex set and $w_{ij}\ge 0$ the weight of the ordered pair $(i,j)$, with $w_{ii}=0$. A binary $x_i$ assigns vertex $i$ to one side of the cut, and a binary $y_{ij}$ may equal one only when the pair $(i,j)$ is cut,
\begin{align}
\max_{x,\,y}\quad & \sum_{i\in V}\sum_{j\in V} w_{ij}\, y_{ij} \label{eq:app-maxcut}\\
\text{s.t.}\quad & y_{ij} - x_i - x_j \le 0, && i,j\in V, \tag{cut\_a}\\
& y_{ij} + x_i + x_j \le 2, && i,j\in V, \tag{cut\_b}\\
& x_i\in\{0,1\},\quad y_{ij}\in\{0,1\}, && i,j\in V. \nonumber
\end{align}
Row (cut\_a) sets $y_{ij}=0$ when both endpoints have $x=0$, and row (cut\_b) sets $y_{ij}=0$ when both have $x=1$.
For $w_{ij}>0$ an optimal solution therefore has $y_{ij} = 1$ exactly when $x_i \neq x_j$. For $w_{ij}=0$, $y_{ij}$ is unconstrained on cut pairs and does not affect the objective. The same two rows force $y_{ii}=0$. The instance has $n=|V|+|V|^2$ binary variables and $m=2|V|^2$ rows, and the pair $\{i,j\}$ enters the objective with effective weight $w_{ij}+w_{ji}$. In the template,

\begin{footnotesize}
\begin{verbatim}
MILP max_cut max
set V 3
par w[V,V]
var x[V] binary
var y[V,V] binary
obj max sum i in V, j in V: w[i,j]*y[i,j]
con cut_a: for i in V, j in V: y[i,j] - x[i] - x[j] <= 0
con cut_b: for i in V, j in V: y[i,j] + x[i] + x[j] <= 2
DATA: {"w":[[0,4,1],[0,0,3],[2,0,0]]}
\end{verbatim}
\end{footnotesize}

\noindent The expansion gives $n=12$ and $m=18$. For the pair $(1,2)$, for instance, it produces $y_{12}-x_1-x_2\le 0$ and $y_{12}+x_1+x_2\le 2$, and for the diagonal entry $(1,1)$ it produces $y_{11}-2x_1\le 0$ and $y_{11}+2x_1\le 2$. The effective weights are $4$ on $\{1,2\}$, $3$ on $\{2,3\}$ and $1+2=3$ on $\{1,3\}$, so the total off-diagonal weight is $W_{\mathrm{off}}=10$. A triangle can cut at most two of its three edges, and the optimum $v^\star=7$ is attained by separating vertex~$2$ from $\{1,3\}$, or equally vertex~$1$ from $\{2,3\}$. The point $x_i=\tfrac12$, $y_{ij}=1$ satisfies every row, and $y_{ij}\le 1$, so the LP relaxation attains $v_{\mathrm{LP}}=\sum_{i,j} w_{ij} = W_{\mathrm{off}} = 10$, using $w_{ii}=0$. The relative relaxation gap $(v_{\mathrm{LP}}-v^\star)/v_{\mathrm{LP}}$ is then $1-s$ with cut share $s=v^\star/W_{\mathrm{off}}=0.7$.

\subsection{Length of the compact and expanded encodings}
\label{app:index-set-length}

Table~\ref{tab:app-length} compares our template with a fully expanded encoding that spells out every objective term, constraint row, and variable declaration using short identifiers (e.g., \texttt{x\_3\_7}, \texttt{y\_2\_5}), evaluated on the training geometries in Appendix~\ref{app:geometries}. For each instance, we sample data uniformly from the prompt ranges and in the max-cut weight matrix, roughly half of the off-diagonal entries are zero. We report lengths in characters. Because the structure block is fixed within each problem family, the compact encoding grows only through its data block. For these dense families the data block and the number of nonzeros in $A$ are of the same order, $O(|F||D|)$ and $O(|V|^2)$ respectively, so the saving is a constant factor, driven by writing each parameter value once instead of spelling out identifiers and operators for every nonzero.

\begin{table}[h]
\centering
\small
\caption{\textbf{Encoding length in characters:} $n$ and $m$ are the numbers of variables and constraints after expansion. Compact is the index-set template including its data block, and Expanded writes every constraint explicitly. Ratio is Compact divided by Expanded.}
\label{tab:app-length}
\begin{tabular}{@{}lrrrrrr@{}}
\toprule
Family & Geometry & $n$ & $m$ & Compact & Expanded & Ratio \\
\midrule
Facility location & $|F|{=}8,\ |D|{=}12$  & 104 & 116 & 786 & 7{,}827  & 0.10 \\
Facility location & $|F|{=}10,\ |D|{=}15$ & 160 & 175 & 962 & 12{,}279 & 0.08 \\
Max-cut           & $|V|{=}9$             & 90  & 162 & 438 & 6{,}401  & 0.07 \\
Max-cut           & $|V|{=}12$            & 156 & 288 & 585 & 11{,}727 & 0.05 \\
\bottomrule
\end{tabular}
\end{table}


\section{Implementation Details}
\label{app:impl}

\subsection{Size curriculum}
\label{sec:curriculum}

We use instance size (number of variables) as the curriculum axis. The direct arms (\gdirect{}, \qdirect{}) train only on the target bracket, $171$--$225$ variables. The curriculum arms (\gladder{}, \qladder{}) move through three rungs, $76$--$110 \rightarrow 111$--$170 \rightarrow 171$--$225$, training for $61$ steps per rung and initializing each rung from the previous rung’s checkpoint.

The smaller rungs are mostly about teaching the model to produce outputs of the right length. As instances get larger, they more often hit the token cap. A truncated instance typically fails to parse and therefore gets a score of zero. When \gbase{} is trained directly on the largest bracket, the valid share drops from over $90\%$ to under $60\%$ and training becomes unstable, so we stop that run after $47$ of $61$ updates (Appendix~\ref{app:res-training}). Warm-starting each rung from a model that already produces complete instances at a slightly smaller size largely avoids this failure mode. By contrast, both Qwen arms stay near $100\%$ validity throughout, so the curriculum is important for Gemma but not for Qwen.

We also need each rung to contain instances that actually branch, otherwise the node-based reward term is identically zero and GRPO gets no hardness signal. For each rung, we therefore screen candidate index-set sizes with the training solver and keep those whose instances branch at least as often as the target-bracket configuration (Appendix~\ref{app:geometries}). In this screen, CFL tends to branch when facilities are few and customers are many, while multiple knapsack never branches at $76$--$110$ variables. As a result, the first rung trains only CFL and Max-Cut, and knapsack is introduced at the second rung.

\subsection{Training configuration}
\label{sec:training-config}

Each arm starts from its model’s instruction-tuned checkpoint and is fine-tuned with LoRA. We compute GRPO statistics within a single problem family and size bracket, alternating families across steps. All other hyperparameters are held fixed across arms and rungs (Table~\ref{tab:hyper}); the only difference is the size-bracket schedule.

\begin{table}[h]
\centering
\small
\caption{\textbf{Training and reward configuration.} Settings are fixed across arms and rungs. Only the size-bracket schedule differs.}
\label{tab:hyper}
\begin{tabular}{@{}llll@{}}
\toprule
\multicolumn{2}{@{}l}{\emph{Policy and optimization}} & \multicolumn{2}{l}{\emph{Reward and solver}} \\
\midrule
Base models           & Gemma-4-12B-it, Qwen3.5-4B     & $N_{\mathrm{ref}}$              & $5{,}000$ \\
Adaptation            & LoRA, rank $16$                & Node cap                        & $50{,}000$ \\
Algorithm             & GRPO, group $G = 64$           & Wall-clock kill (training)      & $20$\,s \\
Learning rate         & $5\times 10^{-5}$              & In-loop solver                  & SCIP 10.0 \\
KL coefficient $\beta$& $0.1$                          & Evaluation-only solvers         & HiGHS, Gurobi 13.0.3 \\
Decoding              & $T=1.0$, top-$p=0.95$    & & \\
Steps per rung        & $61$ & & \\
\bottomrule
\end{tabular}
\end{table}

\subsection{Measurement}
\label{app:measurement}

We proxy instance hardness using (i) the branch-and-bound node count and (ii) the post-root-cut optimality gap $|z^\star - z_{\mathrm{cut}}|/|z^\star|$. We do not use wall-clock solve time because it is machine-dependent. Within each model, all arms are run on the same prompts with matched random seeds, so we make paired comparisons seed-by-seed. For each table cell we generate completions from three prompts that are identical except for the in-context exemplar. This tests sampling \emph{depth} over exemplars rather than \emph{breadth} over many distinct prompts. Each Gemma cell contains $256$ completions and each Qwen cell contains $128$.

If an instance hits the node cap, we keep it and record the node count at the cap. Consequently, high-percentile node statistics should be interpreted as lower bounds. If a solve is terminated by the wall-clock limit, we exclude it from the node and gap summaries, but we still count it in the denominators for parse and feasible rates. The difference between these rates therefore equals the share of killed runs. The default evaluation limit is $300$\,s for two smaller brackets and $3{,}600$ for larger two brackets and for the language-control study.

\section{Rung geometries}
\label{app:geometries}

\paragraph{What we mean by a geometry:} A curriculum rung sets a target variable-count bracket, but an instance is determined by index-set sizes (facilities/customers, vertices, items/knapsacks). For each family and bracket we pick one fixed \emph{geometry} (index-set cardinalities), which uniquely fixes the number of variables $n$ (e.g., facility location: $n=|F|(1+|D|)$; max-cut: $n=|V|+|V|^{2}$; knapsack: $n=|I||K|$). We pass this to the model as a single sentence specifying the bracket and a plausible shape (e.g., ``111--170 variables; about 10 depots and 15 customers''). The in-context exemplar is always shown at a fixed default geometry (facility location $n=150$, max-cut $n=182$, knapsack $n=192$). At small rungs it can be larger than the requested bracket, so the prompt explicitly tells the model to follow the instructions, not the exemplar.

\paragraph{Why we screen geometries:} Some shapes are uninformative. If most instances solve at the root, the node-based reward is near zero and GRPO sees little within-group variation. We therefore want geometries that are not only hard, but also have noticeable spread.

\paragraph{Screening protocol:} Before training (no LM involved), we screened candidate geometries using the family’s reference instance builder. For each geometry we generated 12 randomized instances and solved them with SCIP under the training configuration (node cap $5\times 10^{4}$, 20\,s time limit, $N_{\mathrm{ref}}=5000$ in $H$ \eqref{eq:hard}). All 336 runs finished. This is only a sanity check. If correctly built instances never branch, model-written ones may not either. Each family already had a pre-curriculum ``target-bracket'' geometry (171--225 variables) and we treat this as an \emph{anchor} baseline. We kept a candidate geometry only if (i) at least as many of its 12 instances branched as the anchor (facility location/max-cut: 100\%; knapsack: 75\%), and (ii) the standard deviation of $H$ across the 12 instances exceeded 0.02. Among those that passed, we kept (per family and bracket) the geometry with the largest $\sigma_H$.

\paragraph{Outcome:} Table~\ref{tab:geometries} lists all 28 rows (3 anchors, 25 candidates). Seven candidates passed and five were kept: facility location $|F|{=}8,|D|{=}12$ ($n{=}104$) and $|F|{=}10,|D|{=}15$ ($n{=}160$); max-cut $|V|{=}9$ ($n{=}90$) and $|V|{=}12$ ($n{=}156$); knapsack $|I|{=}42,|K|{=}4$ ($n{=}168$). For max-cut, the two non-kept passing candidates lost a near-tie (e.g., 0.112 vs.\ 0.110 in the 76--110 bracket), which 12 instances can barely separate. For facility location, the branching-rate check was the bottleneck and some rejected shapes had higher median $H$ but branched on only 58--92\% of instances. No knapsack geometry in the 76--110-variable bracket passed. Across eight candidates with $|K|=2$ to 5, at most half of the instances branched (vs.\ the anchor's 75\%), and six of the eight have median $H$ exactly zero. This means the median instance is solved at the root with no post-cut gap. Knapsack is therefore absent from the first bracket training and only enters at the second. Since the first rung trains facility location and max-cut only, its curriculum effect is not fully separable from the change in family mix.

\begin{table}[t]
\centering
\footnotesize
\setlength{\tabcolsep}{4pt}
\caption{\textbf{Screened rung geometries:} Each row is $12$ constructed instances solved by SCIP in the training configuration. $n$ is the variable count, which equals the realized count in every row. Nodes and $H$ are medians over the $12$ instances, $\sigma_H$ is the standard deviation of $H$, and $>$1 node is the share of instances that branch. Status A marks the family's target-bracket anchor, which sets the branching bar. \cmark\ marks the kept geometry, $\circ$ a candidate that passed both conditions but lost the tie-break, and a blank a rejected candidate.}
\label{tab:geometries}
\begin{tabular}{@{}llrcrrrrc@{}}
\toprule
Family & Geometry & $n$ & Bracket & Nodes & $H$ & $\sigma_H$ & $>$1 node (\%) & Status \\
\midrule
\multirow{9}{*}{\shortstack[l]{Facility\\location}}
 & $|F|{=}14,|D|{=}13$ & 196 & 171--225 & 48.5 & 0.426 & 0.155 & 100.0 & A \\
\cmidrule(lr){2-9}
 & $|F|{=}12,|D|{=}7$  &  96 & 76--110  & 11.0 & 0.165 & 0.205 &  75.0 & \\
 & $|F|{=}10,|D|{=}9$  & 100 & 76--110  &  9.0 & 0.251 & 0.203 &  75.0 & \\
 & $|F|{=}13,|D|{=}7$  & 104 & 76--110  & 10.5 & 0.221 & 0.176 &  58.3 & \\
 & $|F|{=}8,|D|{=}12$  & 104 & 76--110  &  7.0 & 0.173 & 0.124 & 100.0 & \cmark \\
\cmidrule(lr){2-9}
 & $|F|{=}13,|D|{=}10$ & 143 & 111--170 & 38.5 & 0.399 & 0.170 &  91.7 & \\
 & $|F|{=}12,|D|{=}12$ & 156 & 111--170 & 34.5 & 0.341 & 0.155 &  91.7 & \\
 & $|F|{=}10,|D|{=}15$ & 160 & 111--170 & 36.0 & 0.421 & 0.081 & 100.0 & \cmark \\
 & $|F|{=}15,|D|{=}10$ & 165 & 111--170 & 47.0 & 0.376 & 0.213 &  83.3 & \\
\midrule
\multirow{5}{*}{Max-cut}
 & $|V|{=}13$ & 182 & 171--225 & 50.5 & 0.580 & 0.108 & 100.0 & A \\
\cmidrule(lr){2-9}
 & $|V|{=}9$  &  90 & 76--110  &  7.0 & 0.407 & 0.112 & 100.0 & \cmark \\
 & $|V|{=}10$ & 110 & 76--110  & 12.0 & 0.427 & 0.110 & 100.0 & $\circ$ \\
\cmidrule(lr){2-9}
 & $|V|{=}11$ & 132 & 111--170 & 18.0 & 0.504 & 0.086 & 100.0 & $\circ$ \\
 & $|V|{=}12$ & 156 & 111--170 & 30.0 & 0.549 & 0.116 & 100.0 & \cmark \\
\midrule
\multirow{14}{*}{\shortstack[l]{Multiple\\knapsack}}
 & $|I|{=}48,|K|{=}4$ & 192 & 171--225 & 2.0 & 0.061 & 0.058 & 75.0 & A \\
\cmidrule(lr){2-9}
 & $|I|{=}48,|K|{=}2$ &  96 & 76--110 & 1.0 & 0.000 & 0.080 & 25.0 & \\
 & $|I|{=}32,|K|{=}3$ &  96 & 76--110 & 1.0 & 0.000 & 0.030 & 16.7 & \\
 & $|I|{=}24,|K|{=}4$ &  96 & 76--110 & 1.0 & 0.000 & 0.030 & 16.7 & \\
 & $|I|{=}34,|K|{=}3$ & 102 & 76--110 & 1.0 & 0.000 & 0.065 & 41.7 & \\
 & $|I|{=}36,|K|{=}3$ & 108 & 76--110 & 1.5 & 0.031 & 0.129 & 50.0 & \\
 & $|I|{=}27,|K|{=}4$ & 108 & 76--110 & 1.0 & 0.000 & 0.093 & 16.7 & \\
 & $|I|{=}55,|K|{=}2$ & 110 & 76--110 & 1.0 & 0.000 & 0.029 & 33.3 & \\
 & $|I|{=}22,|K|{=}5$ & 110 & 76--110 & 1.0 & 0.000 & 0.017 &  8.3 & \\
\cmidrule(lr){2-9}
 & $|I|{=}36,|K|{=}4$ & 144 & 111--170 & 1.0 & 0.000 & 0.043 & 41.7 & \\
 & $|I|{=}40,|K|{=}4$ & 160 & 111--170 & 1.5 & 0.031 & 0.118 & 50.0 & \\
 & $|I|{=}42,|K|{=}4$ & 168 & 111--170 & 2.0 & 0.061 & 0.131 & 75.0 & \cmark \\
 & $|I|{=}28,|K|{=}6$ & 168 & 111--170 & 1.0 & 0.000 & 0.017 &  8.3 & \\
 & $|I|{=}34,|K|{=}5$ & 170 & 111--170 & 1.0 & 0.000 & 0.182 & 41.7 & \\
\bottomrule
\end{tabular}
\end{table}

\section{The P0 generation prompt}
\label{app:promptP0}


P0 is the prompt used for the training where reward was designed using pre-cut gap.



\subsection{System message}

\begin{lstlisting}[style=promptstyle,caption={P0 system message.},label={lst:p0sys}]
You are OptiScribe, an expert in operations research and mathematical optimization. You create realistic, well-posed MIXED-INTEGER Linear Programming (MILP) problems in a COMPACT index-set format that separates structure (written once) from data (numbers in a JSON block).

You draw from deep knowledge of real-world optimization: manufacturing, logistics, finance, energy systems, healthcare operations, and more. Your problems have realistic coefficients, meaningful constraints, integer/binary decisions, and clear optimization objectives.

## Compact index-set format (MILP)

The compact index-set format describes a Linear Program as a STRUCTURE block (text) plus a DATA block (one JSON line). The constraint matrix is NEVER written out -- it is implied by index patterns and rebuilt by expansion. Write each variable family and constraint family ONCE; never enumerate variables one by one.

STRUCTURE lines (in this order):
  LP <name> <min|max>                      problem class and objective sense
  set <NAME> <size>                         an index set, e.g.  set I 10
  par <name>[<SET>,...]                     a parameter table; its numbers live in DATA
  var <name>[<SET>,...] <lb> <ub> <type>    a variable family over those sets; type = continuous
  obj <min|max> sum <i in S, j in T,...>: <terms>      objective summed over index sets
  con <label>: for <i in S,...>: sum <j in T,...>: <lhs> <op> <rhs>    a CONSTRAINT FAMILY
  con <op> <rhs>: <idx>:<coef> ...          (optional) one explicit row, vars by 0-based index
  <op> is one of  <=  >=  = .  <rhs> is a number or a parameter ref like cap[i].
  Terms look like  c[i,j]*x[i,j]  (parameter times variable) or just  x[i,j].

DATA line:
  DATA: {"<par>": <nested JSON array indexed by its sets, row-major>, ...}

WORKED EXAMPLE -- a 2x3 transportation problem (6 variables):
  LP transport min
  set P 2
  set M 3
  par c[P,M]
  par sup[P]
  par dem[M]
  var x[P,M] 0 inf continuous
  obj min sum p in P, m in M: c[p,m]*x[p,m]
  con supply: for p in P: sum m in M: x[p,m] <= sup[p]
  con demand: for m in M: sum p in P: x[p,m] >= dem[m]
  DATA: {"c":[[4,6,5],[7,3,8]],"sup":[50,60],"dem":[20,30,25]}
This expands to variables x_0_0..x_1_2, two supply rows and three demand rows.

RULES:
- Index variables in a family range over their declared sets; x[i,j] is the variable at position (i,j).
- The NUMBER OF DECISION VARIABLES equals the product of the variable family's set sizes.
- Every parameter referenced must be declared with `par` and given values in DATA, shaped by its sets.
- The problem must be feasible and bounded.

## MILP extension (integer and binary variables)

This is a MIXED-INTEGER linear program. The format is identical to the LP index-set format above,
with ONE extension to the `var` line's type slot, which is now one of `continuous | integer | binary`:

  var x[I,J] 0 inf continuous     a continuous family (as in LP)
  var y[I]   0 5   integer        a general-integer family, explicit bounds [0,5]
  var z[I,J] binary               a BINARY family -- its bounds are IMPLICIT 0/1 (write no bounds)

RULES for integer/binary variables:
- A `binary` variable is ALWAYS in {0,1}. Do NOT write its bounds, and NEVER write a constraint row
  that restates a 0/1 bound (e.g. `z[i] <= 1`); declare it with the `binary` type instead.
- Put a general integer's bounds in its `var` DECLARATION (`var y[I] 0 5 integer`), never as rows.
- Mixed-type linear rows are allowed and encouraged: a big-M linking row couples a continuous
  variable to a binary decision, e.g. `sum j: x[i,j] - 100*z[i] <= 0` forces all `x[i,.]=0` when
  z[i]=0 and permits flow up to the big-M when z[i]=1.

WORKED MILP EXAMPLE -- capacitated fixed-charge (6 continuous flows + 3 binary open/close):
  MILP fixed_charge min
  set I 3
  set J 2
  par f[I]
  par c[I,J]
  par d[J]
  var x[I,J] 0 inf continuous
  var z[I] binary
  obj min sum i in I, j in J: c[i,j]*x[i,j] + sum i in I: f[i]*z[i]
  con demand: for j in J: sum i in I: x[i,j] >= d[j]
  con link:   for i in I: sum j in J: x[i,j] - 100*z[i] <= 0
  DATA: {"f":[50,40,60],"c":[[4,6],[7,3],[5,8]],"d":[8,5]}
This is a GENUINE MILP: its LP relaxation opens facilities fractionally, so the integer optimum
differs from the relaxation (a non-trivial integrality gap). `demand` couples the flow variables;
`link` is a big-M family coupling flow to the open/close binaries; no binary is bounded by a row.

## Output Rules

Output ONLY the formulation: the structure lines, then a single final line that starts with "DATA:" and contains the JSON. Start your output with "MILP ". No explanations, no markdown, no other text.
\end{lstlisting}

\subsection{User message}

Braces mark the three per-prompt quantities (Table~\ref{tab:p0sampled}). Everything else is fixed.

\begin{lstlisting}[style=promptstyle,caption={P0 user message. \texttt{\{exemplar\}} is
the few-shot instance of Listing~\ref{lst:p0shot}.},label={lst:p0user}]
Here are examples of MILP problems in the COMPACT index-set format:

<example_1>
{exemplar}
</example_1>

Generate a NEW, original MIXED-INTEGER LP problem about **{domain}** in the SAME compact index-set format.

SPECIFICATIONS:
- The EXPANDED problem must have about {n} decision variables. Choose index-set sizes whose PRODUCT is about {n} (e.g. for ~100 use two sets of size 10; for ~12 use sizes 3 and 4).
- About {k} of those decision variables should be INTEGER or BINARY (declare them with the `integer`/`binary` type). The rest are continuous.
- It must be a GENUINE MILP: the integer/binary decisions must actually matter, so that the LP relaxation (dropping integrality) is NOT already integral. Use integer/binary variables to model indivisible choices -- open/close, select, count, assign -- not just continuous quantities rounded.
- Use 2 to 4 DISTINCT constraint families with DIFFERENT index patterns. At least TWO must be COUPLING families -- families whose expanded rows each sum over an index set of size >= 2, so every row links MULTIPLE variables (a row-sum family, a column-sum family, a weighted budget/knapsack `sum i: w[i]*z[i] <= C`, or a set-cover/assignment family `sum i: z[i,j] = 1`).
- Demonstrate BIG-M LINKING as a first-class pattern: couple a continuous variable to a binary decision with a row like `for i in I: sum j in J: x[i,j] - M[i]*z[i] <= 0`, where the big-M is a NUMERIC constant (e.g. `- 1000*z[i]`) or an indexed parameter `M[i]` declared with `par M[I]`, so that z[i]=0 shuts the continuous variables off and z[i]=1 permits them up to the big-M.
- Put simple per-variable limits in the `var` DECLARATION as bounds (`var x[I,J] 0 50 continuous`, `var y[I] 0 5 integer`), NOT as constraint rows. NEVER write a constraint row that restates a declared bound. FORBIDDEN: writing a binary's 0/1 bounds as rows (e.g. `z[i] <= 1`) -- declare it with the `binary` type, which makes 0/1 implicit.
- Express the objective and constraints as index-set families (`sum i in S, j in T: ...`, `for i in S: sum j in T: ...`) so the structure is written ONCE -- do NOT enumerate variables individually.
- Put every numeric coefficient in the final DATA line as JSON arrays shaped by the sets.
- Use realistic coefficients for the domain.
- The problem MUST be feasible and bounded.

HOW CONSTRAINT ROWS EXPAND (your output is expanded and solved as a MILP):
- `con lbl: for i in S: sum j in T: ... <op> ...` expands to |S| rows; each row sums |T| variables.
- `con lbl: for i in S, j in T: ... <op> ...` expands to |S|*|T| single-variable rows -- use sparingly, only for genuinely per-cell limits.
- `con lbl: sum i in S, j in T: ... <op> ...` (no `for`) is 1 global row over all |S|*|T| variables.

HOW TO ENSURE FEASIBILITY (your output is expanded and solved by a MIP solver):
Build the numbers around a feasible integer operating point so that adding more constraints never makes the problem infeasible:
1. Put simple bounds in the var DECLARATION; keep binaries as the `binary` type (implicit 0/1).
2. Pick a concrete feasible point: choose which binaries are 1 (e.g. enough facilities open), then a modest continuous/integer allocation consistent with those choices.
3. Set each right-hand side FROM that point with a little slack:
   - for a `<=` row, choose rhs >= (the row's left-hand side at the point);
   - for a `>=` row, choose rhs <= (the row's left-hand side at the point);
   - for an `=` row, choose rhs = (the row's left-hand side at the point).
4. For a big-M row `sum x - M*z <= 0`, pick M at least as large as the largest total the continuous variables can reach, so an open (z=1) facility is not artificially throttled.
5. Do not put contradictory families on the same variables.

Output the structure lines, then a single final line starting with "DATA:". Start with "MILP " -- nothing else before or after.
\end{lstlisting}

\subsection{Few-shot exemplar}
\emph{Every} P0 training prompt at \emph{every} curriculum bracket carried the same 5-variable instance (Listing~\ref{lst:p0shot}).

\begin{lstlisting}[style=promptstyle,caption={The fixed training exemplar: 5 variables, 21 lines.},label={lst:p0shot}]
MILP optmath_milp_86390 max
set S0 4
set S1 1
par c_x1[S1]
var z0[S0] binary
var x1[S1] 0 inf continuous
obj max sum i in S1: c_x1[i]*x1[i]
con = 2: 0:1 1:1 2:1 3:1
con <= 2000012: 4:1 0:1000000 1:1000000
con <= 2000011: 4:1 0:1000000 2:1000000
con <= 2000013: 4:1 0:1000000 3:1000000
con <= 2000013: 4:1 1:1000000 0:1000000
con <= 2000012: 4:1 1:1000000 2:1000000
con <= 2000011: 4:1 1:1000000 3:1000000
con <= 2000012: 4:1 2:1000000 0:1000000
con <= 2000015: 4:1 2:1000000 1:1000000
con <= 2000013: 4:1 2:1000000 3:1000000
con <= 2000011: 4:1 3:1000000 0:1000000
con <= 2000011: 4:1 3:1000000 1:1000000
con <= 2000015: 4:1 3:1000000 2:1000000
DATA: {"c_x1":[1.0]}
\end{lstlisting}

\subsection{Per-prompt sampled quantities}
The integer target is stated as a \textbf{count} (``About $k$ of those decision variables\dots''),
not a percentage. Sampling $\rho$ uniformly and rounding would request fractions that do not exist
at small $n$, at $n=14$ the attainable values are $k/14$, spaced $0.071$ apart, so the model
would be penalised for a rounding error it cannot avoid. Drawing $k$ first removes that.

\begin{center}
\small
\captionof{table}{The three quantities resampled for every P0 prompt. $n$ is drawn uniformly from the
current curriculum bracket; the integer \emph{count} $k$ is drawn first and the fraction derived
as $\rho^\star = k/n$, so that the requested fraction is always attainable at that $n$.}
\label{tab:p0sampled}
\begin{tabular}{lll}
\toprule
Symbol & Meaning & Distribution \\
\midrule
$n$      & target variable count & $\mathcal{U}$(curriculum bracket) \\
$k$      & target integer/binary count & $\mathcal{U}\{\lceil 0.05n\rceil,\dots,\lfloor 0.50n\rfloor\}$ \\
$\rho^\star$ & implied integer fraction & $k/n$ (derived, not sampled) \\
domain   & application domain string & uniform over a fixed pool of 25 domains \\
\bottomrule
\end{tabular}
\end{center}

\section{Additional Results}
\label{app:results}

\subsection{Training dynamics}
\label{app:res-training}
Figure~\ref{fig:training_curve} tracks each reward term during training, with rejected completions scored zero. By construction the diversity term averages exactly $0.5$ over valid completions (Appendix~\ref{app:reward-div}), so its curve is half the valid share. It shows how direct training fails on Gemma. The valid share of \gdirect{} falls from above $90\%$ to below $60\%$, and its node score falls with it. The run became unstable and we stopped it after $47$ updates. \gladder{} dips when it moves to a larger rung and recovers within that rung. Both Qwen models stay near a valid share of $100\%$, including \qdirect{}, which trains on the target size from the start. This is why the curriculum matters for Gemma and not for Qwen.

\begin{center}
    \includegraphics[width=\linewidth]{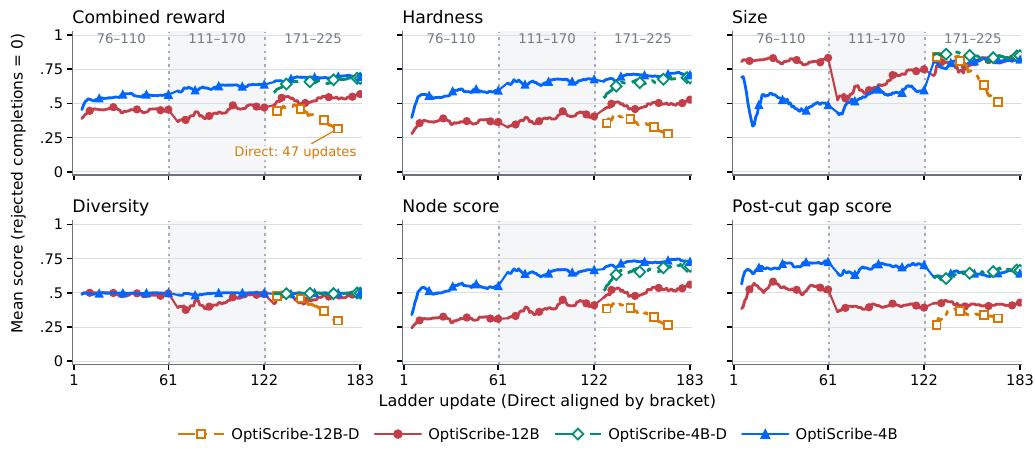}
    \captionof{figure}{\textbf{Training curves:} Mean score per update over all completions, with rejected completions scored $0$. Panels show the combined reward $R$, the hardness $H$, the size term $r_{\mathrm{var}}$, the diversity term $r_{\mathrm{div}}$, the node score $r_{\mathrm{node}}$ and the post-cut gap score $r_{\mathrm{cut}}$. Curriculum models train $61$ updates on each of $76$--$110$, $111$--$170$ and $171$--$225$. Direct models train on $171$--$225$ only and are aligned with the last rung. \gdirect{} became unstable and was stopped after $47$ updates. The diversity term averages exactly $0.5$ over valid completions, so its curve is half the valid share.}
    \label{fig:training_curve}
\end{center}

\subsection{Multiple knapsack}
\label{app:res-knapsack}
Table~\ref{tab:main-g1} reports multiple knapsack, the third training family, at the two brackets on which it was trained. No model makes it hard. Median node counts stay between $3$ and $5$ and median gaps at $4$--$5\times10^{-3}$ for every model. Knapsack also sat out the first curriculum rung because no geometry at $76$--$110$ variables branched (Appendix~\ref{app:geometries}). We regard it as an open case for the method.

\begin{center}
\small
\captionof{table}{\textbf{Multiple knapsack under SCIP at the two trained brackets.} Columns as in Table~\ref{tab:main-f5-g2-split}. No model generates hard knapsack instances, and training leaves node counts and gaps essentially unchanged.}
\label{tab:main-g1}
\begin{tabular}{llcccc}
\toprule
Bracket & Arm & Parse\,$\uparrow$ & Feas\,$\uparrow$ & Nodes\,$\uparrow$ & Gap\,$\uparrow$ \\
\midrule
\multirow{3}{*}{111--170}
 & \gbase{}   & 88.3 & 85.5 & 3 & 5 \\
 & \gdirect{} & 77.0 & 75.4 & 3 & 5 \\
 & \gladder{} & \textbf{93.4} & \textbf{91.4} & \textbf{5} & 5 \\
\cmidrule(lr){1-6}
\multirow{3}{*}{171--225}
 & \gbase{}   & \textbf{91.8} & \textbf{90.2} & \textbf{5} & 4 \\
 & \gdirect{} & 69.1 & 64.5 & 3 & 4 \\
 & \gladder{} & 87.5 & 82.8 & \textbf{5} & 4 \\
\bottomrule
\end{tabular}
\end{center}

\subsection{Standard synthetic generators}
\label{app:res-ecole}
Learning-for-MILP papers often benchmark on four synthetic families, i.e., set cover, independent set, combinatorial auction, and capacitated facility location, available in Ecole \citep{gasse2019exact,prouvost2020ecole}. We ran these generators and solved them with the same SCIP settings as our instances (Table~\ref{tab:app-ecole}). In the $111$--$500$ variable range, $55\%$ solve at the root and none exceeds $189$ nodes. Even at the published sizes (up to $10{,}100$ variables) the median is $21$ nodes and the maximum $853$. Facility location is the closest structural match, but despite a similar integer share (generator: $0.059$, our CFL: $0.063$), \gladder{} needs much more search (median $347$ vs $1$ node), likely because these generators were tuned around SCIP~6 while SCIP~10 presolves and cuts more aggressively.

\begin{center}
\small
\setlength{\tabcolsep}{5pt}
\captionof{table}{\textbf{Standard synthetic generators vs. \gladder{} under SCIP.} Four families from \citet{gasse2019exact} (Ecole; \citealp{prouvost2020ecole}). Top block rescales to $111$--$500$ variables, bottom uses published sizes. \emph{Variables} is the median variable count. \emph{Branching} is the share needing $>1$ node. SCIP uses a $50{,}000$-node cap and a $300$\,s limit. No standard-generator instance hits either. \gladder{} node stats exclude $300$\,s timeouts ($2.6\%$ CFL, $6.4\%$ Max-Cut).}
\label{tab:app-ecole}
\begin{tabular}{@{}lrrrrr@{}}
\toprule
Source & Variables & Nodes p50 & Nodes p90 & Nodes max & Branching (\%) \\
\midrule
\multicolumn{6}{@{}l}{\emph{Standard generators, $111$--$500$ variables, $250$ instances each}} \\
Combinatorial auction & 319 & 4 & 11 & 70 & 78.0 \\
Independent set       & 299 & 1 & 15 & 189 & 48.8 \\
Set cover             & 304 & 1 & 5 & 28 & 33.6 \\
Facility location     & 272 & 1 & 3 & 20 & 18.8 \\
Pooled                & --  & 1 & 8 & 189 & 44.8 \\
\addlinespace
\multicolumn{6}{@{}l}{\emph{Standard generators, published sizes, $25$ instances each}} \\
Combinatorial auction & 500 & 13 & 50 & 120 & 100.0 \\
Independent set       & 500 & 15 & 229 & 853 & 84.0 \\
Set cover             & 1{,}000 & 37 & 387 & 590 & 100.0 \\
Facility location     & 10{,}100 & 136 & 384 & 492 & 100.0 \\
Pooled                & --  & 21 & 271 & 853 & 96.0 \\
\addlinespace
\multicolumn{6}{@{}l}{\emph{\gladder{}, $111$--$500$ variables, standard prompt}} \\
CFL     & 196 & 347 & 4{,}655 & 50,000 & 96.3 \\
Max-Cut & 210 & 634 & 2{,}643 & 14,678 & 100.0 \\
\bottomrule
\end{tabular}
\end{center}

\subsection{Public benchmark sources}
\label{app:public-datasets}
Table~\ref{tab:public-sources} lists the public sources pooled in Section~\ref{sec:res-public}. From each source we keep instances with $111$--$500$ variables. We then split them by variable type: mixed-integer instances are compared against our CFL pools, and pure-integer instances against our Max-Cut pools.

\begin{center}
\small
\captionof{table}{Public MILP instance sources used in Section~\ref{sec:res-public}.}
\label{tab:public-sources}
\begin{tabular}{@{}lp{0.62\linewidth}@{}}
\toprule
Source & Description \\
\midrule
MIPLIB 2010 \citep{koch2011miplib2010} & Fifth edition of the Mixed Integer Programming Library. \\
MIPLIB 2017 \citep{miplib2017} & Sixth edition of the Mixed Integer Programming Library, compiled with a data-driven selection procedure. \\
MILP-Evolve \citep{li2025milpevolve} & MILP problem classes generated by an evolutionary framework based on large language models (LLMs). \\
D-MIPLIB \citep{huang2024distributional} & Distributional MIPLIB: a multi-domain library of MILP instance distributions for machine-learning-guided methods. \\
DIG-MILP \citep{wang2024digmilp} & Instances from the DIG-MILP study, which uses a deep generator (a variational autoencoder, VAE) that guarantees feasibility. \\
MIPcc23 \citep{bolusani2024mipcc23} & MIP Workshop 2023 Computational Competition on reoptimization. Each instance series contains related instances of the same size. \\
MIPLearn \citep{xavier2024miplearn} & Benchmark problems distributed with the MIPLearn framework for learning-enhanced optimization. \\
MIRPLIB \citep{papageorgiou2014mirplib} & Library of maritime inventory routing problem instances. \\
\bottomrule
\end{tabular}
\end{center}

\subsection{Node counts of all models under three solvers}
\label{app:res-solvers}
Table~\ref{tab:app-solvers} extends Figure~\ref{fig:solver_ablation} to all three Gemma models. The ordering of Table~\ref{tab:main-f5-g2-split} holds under each solver. \gladder{} has the highest median in every cell except CFL at $111$--$170$ under Gurobi, where the median instance of every model is solved at the root. On CFL, \gdirect{} is at or below the base model under every solver and in every bracket. Absolute counts differ by solver. Gurobi needs far fewer nodes than SCIP on CFL and more on Max-Cut.

\begin{center}
\small
\captionof{table}{\textbf{Median branch-and-bound nodes under three solvers.} All three Gemma models, both families and all four brackets. SCIP is the training solver. HiGHS and Gurobi never enter training. Max-Cut instances are solved as maximization. Bold marks the highest median in each bracket and solver. \gladder{} is highest in every cell except CFL at $111$--$170$ under Gurobi, where the median instance of every model is solved at the root.}
\label{tab:app-solvers}
\begin{tabular}{llccccccc}
\toprule
& & \multicolumn{3}{c}{CFL} & & \multicolumn{3}{c}{Max-cut)} \\
\cmidrule(lr){3-5}\cmidrule(lr){7-9}
Bracket & Arm & SCIP\,$\uparrow$ & HiGHS\,$\uparrow$ & Gurobi\,$\uparrow$ & &
                SCIP\,$\uparrow$ & HiGHS\,$\uparrow$ & Gurobi\,$\uparrow$ \\
\midrule
\multirow{3}{*}{111--170}
 & \gbase{}   & 22 & 19 & 1 & & 41 & 27 & 10 \\
 & \gdirect{} & 21 & 16 & 1 & & 41 & 29 & 15 \\
 & \gladder{} & \textbf{109} & \textbf{79} & 1 & & \textbf{142} & \textbf{59} & \textbf{214} \\ 
\cmidrule(lr){1-9}
\multirow{3}{*}{171--225}
 & \gbase{}   & 224 & 277 & 58 & & 81 & 41 & 36 \\
 & \gdirect{} & 170 & 179 & 2 & & 85 & 41 & 44 \\
 & \gladder{} & \textbf{391} & \textbf{491} & \textbf{69} & & \textbf{364} & \textbf{158} & \textbf{982} \\
\cmidrule(lr){1-9}
\multirow{3}{*}{226--350}
 & \gbase{}   & 293 & 480 & 53 & & 423 & 200 & 412 \\
 & \gdirect{} & 211 & 194 & 1 & & 369 & 163 & 302 \\
 & \gladder{} & \textbf{710} & \textbf{956} & \textbf{126} & & \textbf{1002} & \textbf{873} & \textbf{4379} \\
\cmidrule(lr){1-9}
\multirow{3}{*}{351--500}
 & \gbase{}   & 420 & 539 & 69 & & 1350 & 1826 & 4363 \\
 & \gdirect{} & 194 & 230 & 1 & & 1096 & 1303 & 3353 \\
 & \gladder{} & \textbf{1014} & \textbf{1539} & \textbf{176} & & \textbf{2578} & \textbf{5041} & \textbf{6101} \\
\bottomrule
\end{tabular}
\end{center}

\subsection{Trivial instances}
\label{app:res-trivial}
Figure~\ref{fig:solver_triviality} reports the share of instances each solver solves at the root node. For \gladder{}, which was trained with SCIP, SCIP and HiGHS solve at most $7\%$ of CFL instances at the root and almost no Max-Cut instances. Gurobi solves $35$--$52\%$ of CFL instances at the root but almost no Max-Cut instances. The pre-cut model, trained with HiGHS, behaves differently. HiGHS solves at most $6\%$ of its instances at the root, while SCIP solves $33$--$57\%$ and Gurobi $38$--$49\%$. Its instances are still solved at the root less often than the base model's on the same prompt, which SCIP and Gurobi solve at the root $85$--$98\%$ of the time.

\begin{center}
    \includegraphics[width=\linewidth]{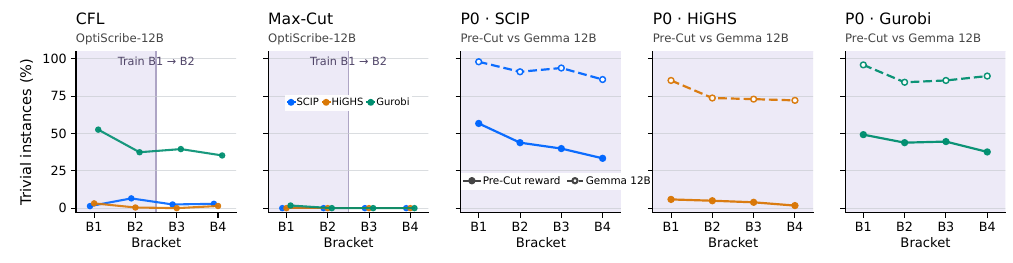}
    \captionof{figure}{\textbf{Share of instances solved at the root under three solvers.} The two left panels show \gladder{} on CFL and Max-Cut, with shading on the training brackets. The three right panels show prompt P0 under SCIP, HiGHS and Gurobi, for the pre-cut model (solid) and untrained \gbase{} (dashed). The pre-cut model was trained with HiGHS in the loop. HiGHS rarely solves its instances at the root, but SCIP and Gurobi often do.}
    \label{fig:solver_triviality}
\end{center}

\subsection{Sensitivity to the node reference}
\label{app:res-nref}
Figure~\ref{fig:nref} compares two \qladder{} runs that differ in $N_{\mathrm{ref}}$ and in the training node cap. Inside the training range their node and gap gains are similar. Beyond it they differ on nodes and parsing but not on the gap. On CFL, $N_{\mathrm{ref}} = 50{,}000$ gives larger node gains at B3 and B4 ($24.5\times$ and $5.7\times$ against $7.6\times$ and $1.6\times$). On Max-Cut it gives a smaller node gain at B4 ($6.6\times$ against $11.6\times$), and its parse rate falls $13$--$15$ points below the base model at B3 and B4. Post-cut gap gains stay close to each other at every size. Neither setting is better on every metric.

\begin{center}
    \includegraphics[width=\linewidth]{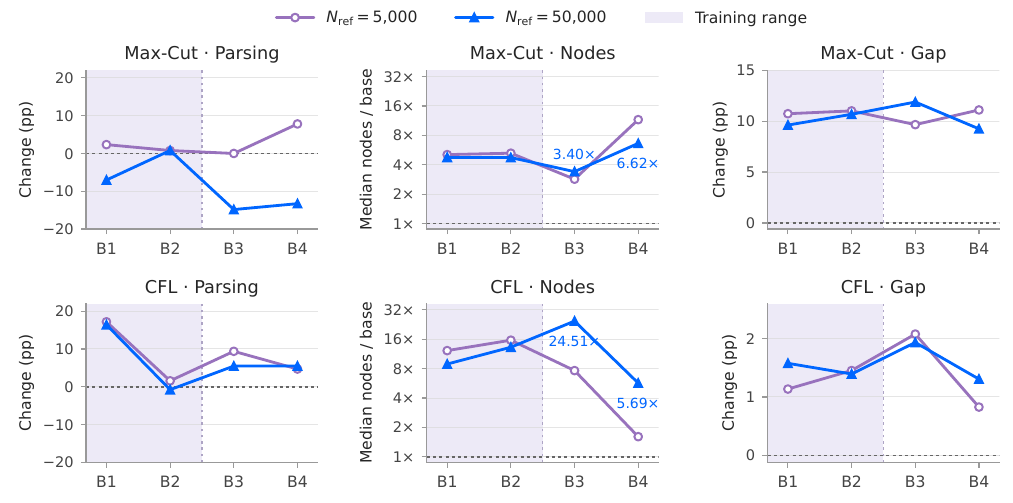}
    \captionof{figure}{\textbf{Sensitivity to the node reference $N_{\mathrm{ref}}$ for \qladder{}.} Each panel shows the change against untrained \qbase{} for a model trained with $N_{\mathrm{ref}}=5{,}000$, the default, and one trained with $N_{\mathrm{ref}}=50{,}000$. The top row is Max-Cut and the bottom row is CFL. Parsing and Gap are changes in percentage points, where Gap is the median post-root-cut gap. Nodes is the ratio of median node counts. Shading marks the training brackets. The two runs also differ in their training node cap.}
    \label{fig:nref}
\end{center}

\subsection{Language control details}
\label{app:res-language}
Table~\ref{tab:sentences} lists the prompt conditions. On Max-Cut the density request is a reliable dial. At requested densities of $0.5$ and $0.2$, every seed of both models lowers its density against the same seed under the control prompt (Figure~\ref{fig:sparsity_effect}b). At $0.8$ the change is small, and $73\%$ of base seeds and $83\%$ of \gladder{} seeds move by at least $0.02$. Sparse instances also solve roughly ten times faster for both models (Figure~\ref{fig:sparsity_effect}a).

On CFL the density sentence asks that only a fraction of facilities be economical for each customer. The base model follows it, with median $\kappa$ of $0.80$, $0.53$ and $0.07$ for requests of $0.8$, $0.5$ and $0.2$. \gladder{} does not, giving $0.97$, $0.84$ and $0.88$. It keeps serving costs close together, which is the opposite of what the sentence asks. Its instances still become easy under this sentence, with median nodes falling from $203$ to between $1$ and $5$, through a change we have not identified. On CFL the word ``harder'' raises median nodes for both models, from $142$ to $215$ and from $203$ to $464$, but with $48$ completions neither paired change is significant.

Instructions cost the trained model less validity. Appending any sentence lowers the base model's CFL parse rate from $94\%$ to as low as $67\%$. For \gladder{} it falls from $98\%$ to no lower than $90\%$. On Max-Cut both models parse every completion under every sentence. \gladder{} writes the complete header in all of them, against $40$--$85\%$ for the base model.

\begin{center}
\footnotesize
\setlength{\tabcolsep}{4pt}
\captionof{table}{\textbf{Prompt conditions for the language-control study.} Each condition appends one sentence to the production prompt at a fixed position. Everything else is byte-identical across conditions and models. This includes the row form, the size request, the exemplar pool, the token cap, the decoding settings and the seed of each completion. $\kappa$ is the coupling density. On Max-Cut it is the fraction of vertex pairs with nonzero effective weight $w_{ij}+w_{ji}$. On CFL it is the mean fraction of facilities whose serving cost is within $10\times$ of each customer's cheapest option. $\kappa^{\mathrm{req}}$ is the value the sentence asks for.}
\label{tab:sentences}
\begin{tabular}{@{}llp{0.55\textwidth}@{}}
\toprule
Tag & $\kappa^{\mathrm{req}}$ & Sentence appended to the production prompt \\
\midrule
control & -- & (nothing appended) \\
harder & -- & Make this instance as hard as possible for a branch-and-bound solver to prove optimal. \\
easier & -- & Make this instance easy, a warm-up example that a solver should finish almost immediately. \\
\addlinespace
\multicolumn{3}{@{}l}{\emph{Max-Cut}} \\
sparser & $0.8/0.5/0.2$ & Use a graph in which about $P$ percent of the vertex pairs are joined by an edge; every remaining pair must have weight zero. \\
sparser $+$ harder & $0.5$ & The sentence above at $P=50$, followed by the difficulty sentence. \\
\addlinespace
\multicolumn{3}{@{}l}{\emph{CFL}} \\
sparser & $0.8/0.5/0.2$ & For each customer only about $P$ percent of the facilities should be economical; every other facility must cost that customer at least ten times its cheapest option. \\
sparser $+$ harder & $0.5$ & The sentence above at $P=50$, followed by the difficulty sentence. \\
\bottomrule
\end{tabular}
\end{center}

\begin{center}
    \includegraphics[width=0.95\linewidth]{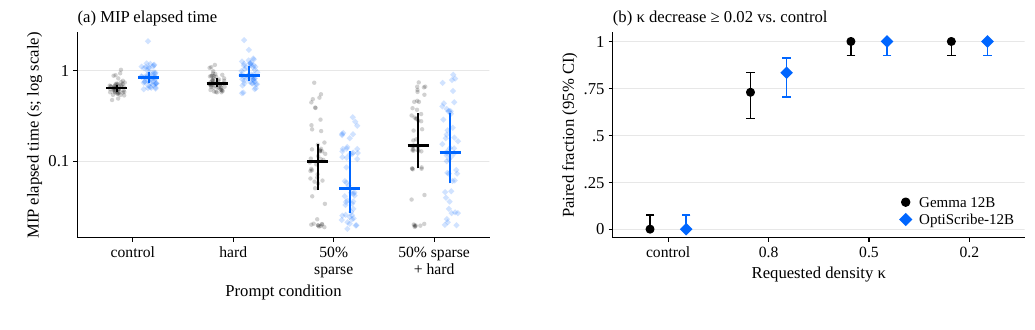}
    \captionof{figure}{\textbf{More on the Max-Cut density instruction.} Bracket $171$--$225$, $48$ completions per model and prompt. (a) SCIP solve time per instance on a log scale. Sparse requests cut solve time by roughly an order of magnitude for both models. (b) Fraction of seeds whose density falls by at least $0.02$ against the same seed under the control prompt, with $95\%$ confidence intervals. Every seed moves at requested densities of $0.5$ and $0.2$.}
    \label{fig:sparsity_effect}
\end{center}

\subsection{Instance properties by bracket}
\label{app:res-properties}
Figures~\ref{fig:features_f5} and~\ref{fig:features_g2} break the results down by bracket for all six models. Three points matter for reading the main tables. First, feasibility given a parse stays between about $86\%$ and $100\%$ for every model, so validity differences come mostly from parsing. Second, constraint-matrix density falls with size at the same rate for every model, so trained models do not buy hardness with denser matrices. Third, size targeting weakens at the largest bracket. At $351$--$500$, \gladder{} lands $56\%$ of CFL and $75\%$ of Max-Cut instances inside the requested bracket. \qladder{} misses the Max-Cut bracket often at every size, landing inside it for only $18$--$71\%$ of instances.

\begin{center}
    \includegraphics[width=0.95\linewidth]{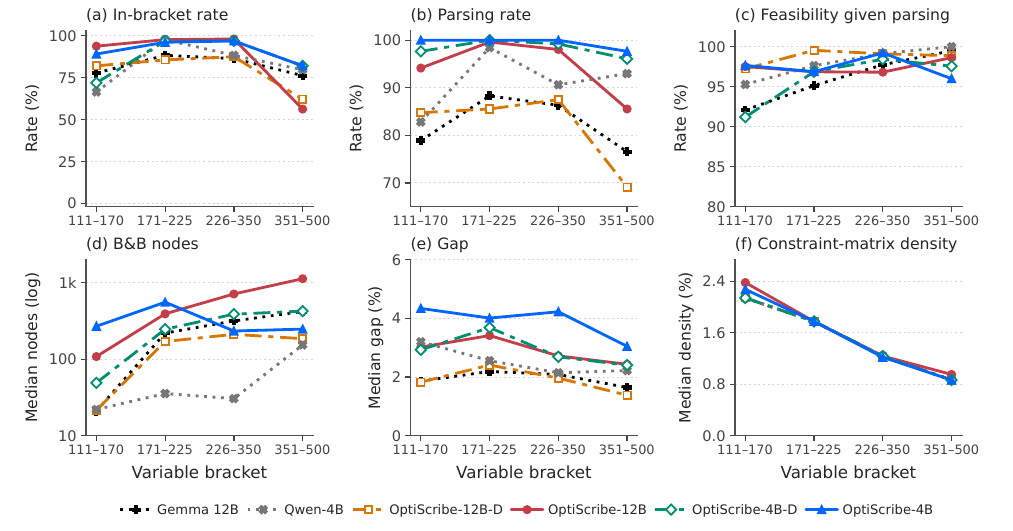}
   \captionof{figure}{\textbf{CFL instance properties by size bracket under SCIP, for all six models.} (a) Share of instances whose variable count falls inside the requested bracket. (b) Parse rate. (c) Feasible rate among parsed completions. (d) Median branch-and-bound nodes. (e) Median post-root-cut gap in percent. (f) Median constraint-matrix density, the share of nonzero entries. Curriculum models are trained on $76$--$225$ variables and direct models on $171$--$225$.}
    \label{fig:features_f5}
\end{center}

\begin{center}
    \includegraphics[width=0.95\linewidth]{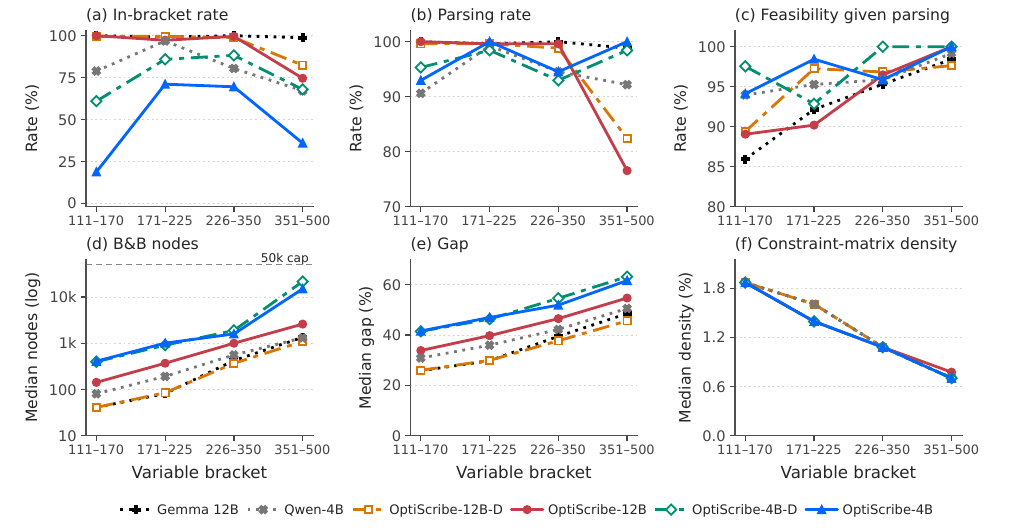}
   \captionof{figure}{\textbf{Max-Cut instance properties by size bracket under SCIP, for all six models.} Panels as in Figure~\ref{fig:features_f5}. The dashed line in (d) is the $50{,}000$ node cap.}
    \label{fig:features_g2}
\end{center}

\subsection{The sense token on Max-Cut}
\label{app:res-sense}
The Max-Cut header is \texttt{MILP max\_cut max}. Untrained \gbase{} writes the final sense token in only $48$--$69\%$ of completions, against $94$--$98\%$ for \gdirect{} and $100\%$ for \gladder{}. Without the token our parser reads the objective as a minimization, whose optimum is zero and is found at the root. The prompt always asks for maximization, so in evaluation we solve every Max-Cut instance as a maximization. The rule applies to all models alike, so the Max-Cut gains in Table~\ref{tab:main-f5-g2-split} do not come from the header. We suspect the base model drops the token because the family name already contains the word max, but we have not tested this.

\end{document}

%% file: math_commands.tex
\usepackage{amsmath,amsfonts,bm}

\def\eqref#1{equation~\ref{#1}}

\def\1{\bm{1}}

\DeclareMathAlphabet{\mathsfit}{\encodingdefault}{\sfdefault}{m}{sl}
\SetMathAlphabet{\mathsfit}{bold}{\encodingdefault}{\sfdefault}{bx}{n}

